\documentclass[manuscript,screen]{style-telo/acmart}

\makeatletter
\def\@ACM@checkaffil{
    \if@ACM@instpresent\else
    \ClassWarningNoLine{\@classname}{No institution present for an affiliation}%
    \fi
    \if@ACM@citypresent\else
    \ClassWarningNoLine{\@classname}{No city present for an affiliation}%
    \fi
    \if@ACM@countrypresent\else
        \ClassWarningNoLine{\@classname}{No country present for an affiliation}%
    \fi
}
\makeatother

\usepackage{listings}

\usepackage{amsmath}
\usepackage{csquotes}
\PassOptionsToPackage{usenames,dvipsnames, table}{xcolor}
\usepackage{xcolor}
\usepackage{tcolorbox}
\usepackage{subcaption}

\usepackage{algorithm}
\usepackage{algpseudocode}

\usepackage{wrapfig}
\usepackage[utf8]{inputenc}
\usepackage{longtable}
\usepackage{array} 

\definecolor{DAEColor}{rgb}{0.1216, 0.4667, 0.7059} 
\definecolor{AEColor}{rgb}{0.1725, 0.6275, 0.1725} 
\definecolor{VAEColor}{rgb}{0.8392, 0.1529, 0.1569} 
\definecolor{LLMColor}{rgb}{1.0000, 0.4980, 0.0549} 

\definecolor{SolutionEncodingColor}{rgb}{0.5804, 0.4039, 0.7412} 
\definecolor{SolutionGenerationColor}{rgb}{0.5490, 0.3373, 0.2941} 
\definecolor{MutationEncodingColor}{rgb}{0.8902, 0.4667, 0.7608} 
\definecolor{InstructedMutationColor}{rgb}{0.7373, 0.7412, 0.1333} 

\definecolor{CurrentRunColor}{rgb}{0.2196, 0.4235, 0.6902} 
\definecolor{PreviousRunsColor}{rgb}{0.6196, 0.8549, 0.8980} 
\definecolor{FoundationModelColor}{rgb}{0.1922, 0.6392, 0.3294} 
\definecolor{FoundationModelCurrentRunColor}{rgb}{0.6431, 0.5686, 0.2706} 

\DeclareMathOperator*{\argmax}{argmax}

\usepackage{amsthm}

\AtBeginDocument{%
  \providecommand\BibTeX{{%
    \normalfont B\kern-0.5em{\scshape i\kern-0.25em b}\kern-0.8em\TeX}}}

\setcopyright{acmcopyright}

\begin{document}

\title[Language Model Crossover: Variation through Few-Shot Prompting]{Language Model Crossover:\\Variation through Few-Shot Prompting}


\author{Elliot Meyerson}
\orcid{0000-0002-1871-2757}
\affiliation{%
  \institution{Cognizant AI Labs}
}
\email{elliot.meyerson@cognizant.com}

\author{Mark J. Nelson}
\orcid{0000-0003-1882-8896}
\affiliation{%
  \institution{American University}
}
\email{mnelson@american.edu}

\author{Herbie Bradley}
\orcid{0000-0001-5390-1257}
\affiliation{%
  \institution{University of Cambridge \& CarperAI}
}
\email{hb574@cam.ac.uk}

\author{Adam Gaier}
\orcid{0000-0002-4632-0929}
\affiliation{%
 \institution{Autodesk Research}
 }
 \email{adam.gaier@autodesk.com}

\author{Arash Moradi}
\orcid{0000-0002-6791-4988}
\affiliation{%
 \institution{New Jersey Institute of Technology}
 }
 \email{am3493@njit.edu}

\author{Amy K. Hoover}
\orcid{0000-0002-4661-8178}
\affiliation{%
 \institution{New Jersey Institute of Technology}
 }
 \email{ahoover@njit.edu}

\author{Joel Lehman}
\orcid{0000-0002-9535-1123}
\affiliation{%
CarperAI}
\email{lehman.154@gmail.com}

\renewcommand{\shortauthors}{Meyerson, Nelson, Bradley, Gaier, Moradi, Hoover and Lehman}

\begin{abstract}
This paper pursues the insight that language models naturally enable
an intelligent variation operator similar in spirit to evolutionary crossover.
In particular, language models of sufficient scale demonstrate in-context learning, i.e.\ they
can learn from associations between a small number of input patterns to generate outputs
incorporating such associations (also called few-shot prompting). This ability can be
leveraged to form a simple but powerful variation operator, i.e.\ to prompt a language
model with a few text-based genotypes (such as code, plain-text sentences, or equations), and
to parse its corresponding output as those genotypes' offspring. The promise of such language model crossover 
(which is simple to implement and can leverage many different open-source language models) is that it
enables a simple mechanism to evolve semantically-rich text representations 
(with few domain-specific tweaks), and naturally benefits from current progress in language models.
Experiments in this paper highlight the versatility of language-model crossover, through evolving
binary bit-strings, sentences, equations, text-to-image prompts, and Python code. The conclusion
is that language model crossover is a flexible and effective method for evolving genomes representable as text.
\end{abstract}




\begin{teaserfigure}
  \includegraphics[width=\textwidth]{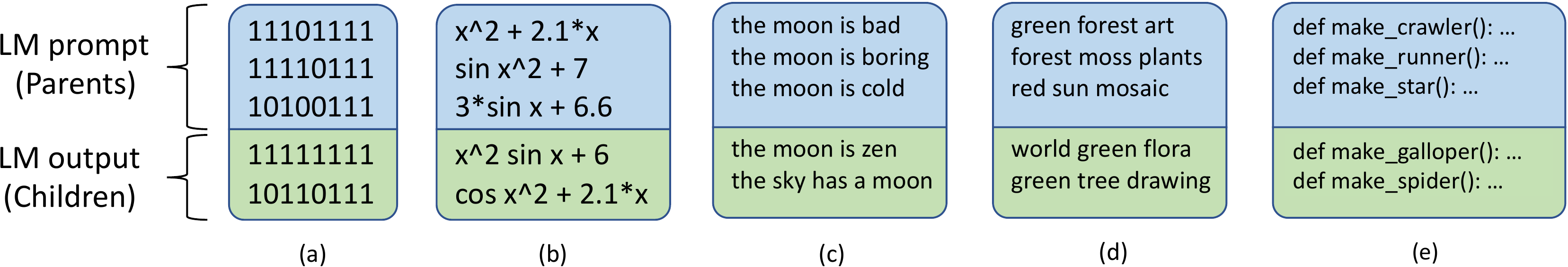}
  \caption{\emph{Language Model Crossover (LMX).}
  New candidate solutions are generated by concatenating parents into a prompt, feeding the prompt through any large pre-trained large language model (LLM), and collecting offspring from the output. Such an operator can be created through very few lines of code. 
  The enormity and breadth of the dataset on which the LLM was trained, along with its ability to perform in-context learning, enables LMX to generate high-quality offspring across a broad range of domains.
  Domains demonstrated in this paper include (a) binary strings, (b) mathematical expressions, (c) English sentences, (d) image generation prompts, and (e) Python code; many more are possible.
  When integrated into an optimization loop, LMX serves as a general and effective engine of text-representation evolution.
  }
  \label{fig:teaser}
\end{teaserfigure}

\maketitle

\section{Introduction}

Large language models (LLMs; \cite{brown2020language,bommasani2021opportunities}) are behind many of the approaches achieving state-of-the-art results in natural language processing domains, such as question-answering \cite{lu:neurips22,cheng:acl21,fajcik:acl21}, code-generation \cite{chen:arxiv21,li:science22}, and few-shot classification \cite{brown2020language,schick:ecacl21}. One popular type of LLM is trained on corpora of human-authored text to predict the next token from previous ones, i.e.\ autoregressive LLMs (e.g.\ GPT-3), which at their core model a distribution of likely output sequences given an input sequence or \emph{prompt}. In zero-shot prompting, a LLM generates an output response from a single input query. However, another popular prompting paradigm is \emph{few-shot prompting} \cite{brown2020language}, wherein the input to the LLM contains a few examples of desired input-output behavior (e.g.\ how to classify a sentence's sentiment) preceding a new target input that the model is to classify. In this way, to some extent such LLMs have \emph{meta-learned} how to learn a desired task given only a few natural-language examples \cite{von2022transformers,chan2022data}. 

One reason this ability is exciting is because it highlights how LLMs can in effect be seen as powerful pattern-completion engines. Few-shot prompting works because the LLM can ``guess the pattern'' behind a few input/output pairs and generalize its behavior to a new target input (provided at the end of the few-shot prompt).
The central insight of this paper is is that the pattern-completion ability of few-shot prompting can be leveraged to create a form of intelligent evolutionary crossover.

For example, if three text-based genotypes are drawn
from a population and concatenated into a prompt, an ideal pattern-completion engine would analyze their commonalities and generate a new (fourth) genotype that qualitatively follows from the same distribution. In effect such an operator would combine aspects of the input genotypes, and indeed, an experiment in Section \ref{sec:binary} demonstrates empirically that LLMs enable this with binary strings. Theoretically we also connect this form of \emph{LLM crossover} (LMX) to estimation of distribution algorithms (EDAs; \cite{baluja:94,larranaga:02}), wherein LMX can be seen
as building an implicit probabilistic model of the input parent genotypes from which to sample a new offspring, \emph{through a single forward pass of the LLM}. From the perspective of intelligent pattern-completion, this operator should naturally improve as LLMs increase in capabilities (which experiments here validate); furthermore, to increase performance the method can easily leverage the rise of open-source domain-specific LLMs that match a target domain (e.g.\ LLMs that focus on code, when the target domain is to evolve code), often with changing only a single line of code to rely on a different hosted model (e.g.\ through the HuggingFace model repository \cite{wolf2019huggingface}).

The benefit of LMX is that evolution can easily and effectively leverage the semantically-rich (and generic) representation of
text, e.g.\ without having to design domain-specific variation operators. LMX's versatility is 
highlighted in experiments with binary strings, style transfer of plain-text sentences, symbolic regression of
mathematical expressions, generating images through prompts for a text-to-image model, and generating Python code. The results highlight the potential of the method to produce quality results across domains,
often by leveraging the broad ecosystem of pretrained models that can be easily combined in many ways to quantify fitness or diversity, or to cross modalities (i.e.\ from text to image).
LMX may also synergize with recent LLM-based mutation techniques \cite{lehman2023evolution}, and is amenable to similar possibilities such as fine-tuning an LLM as a way of accelerating search, although we leave these possibilities for future work (See Section~\ref{sec:discussion}).

In short, the main contributions of this paper are to introduce LMX, explore its basic properties, and highlight its versatility through testing it in a variety of domains. We will release an implementation of LMX and code to recreate the main experiments of the paper.

\section{Background}

This section reviews foundation models and intelligent variation in evolutionary computation.

\subsection{Foundation Models}

A recent paradigm in ML is to train increasingly large models on internet-scale data, e.g.\ BERT and GPT-3 on text \cite{brown2020language,devlin2018bert}, or DALL-E and stable diffusion on captioned images \cite{latentdiffusion:cvpr2022,ramesh2021zero}. Such models are sometimes called foundation models \cite{bommasani2021opportunities}, as they provide a broad foundation from which they can be specialized to many specific domains (e.g.\ with supervised fine-tuning (i.e., further training on a domain-specific dataset) or prompt-engineering). Foundation models have enabled a vibrant ecosystem of specialized models \cite{von2022evaluate} that can be combined in a plug-and-play way (e.g.\ models that measure sentiment of text \cite{camacho:arxiv22tweetnlp}, summarize text \cite{stiennon2020learning}, write code \cite{nijkamp2022codegen}, rank the aesthetics of images \cite{kong:eccv16,deng:ieeespm17,laion-aesthetics}, and create high-dimensional embeddings of text or images \cite{yu:tml22,reimers-2019-sentence-bert}. One contribution of this paper is to demonstrate how evolutionary methods can easily leverage this growing ecosystem to evolve high-quality artifacts in diverse applications.

One particularly exciting class of foundation models are pre-trained language models (LMs) that model the distribution of text. While early LMs used markov chains \cite{shannon2001mathematical} or recurrent neural networks \cite{graves2013generating}, more recently the transformer architecture \cite{vaswani2017attention} has enabled significant progress in NLP. 
Let $V$ be a vocabulary of text tokens, e.g., words or other atomic pieces of text.
Then, $V^*$ is the set of strings made up of tokens from $V$.
Given an input string $a_1 a_2\ldots a_{T_\textrm{in}} \in V^*$, a large autoregressive transformer-based LM (LLM) probabilistically generates an output string:
\begin{equation}
    \label{eq:llm}
    a_{T_\textrm{in} + 1}a_{T_\textrm{in} + 2}\ldots a_{T_\textrm{in} + T_\textrm{out}} \sim \textrm{LLM}(a_1 a_2\ldots a_{T_\textrm{in}}).
\end{equation}
where $a_{T_\textrm{in} + i}$ are all sampled autoregressively:
\begin{equation}
    \label{eq:llm_autoregression}
    a_{T_\textrm{in} + i} \sim \textrm{LLM}_o(a_1 a_2\ldots a_{T_\textrm{in} + i -1}) \ \forall \ i \in [1, T_\textrm{out}],
\end{equation}
where LLM$_o$ is the softmax distribution over $V$ induced by a single forward pass through the transformer model.
The method in this paper focuses on one emergent capability of LLMs: the potential to learn from text examples provided as input to the model when generating an output, which is called \emph{in-context learning} or \emph{few-shot prompting} \cite{brown2020language,von2022transformers}. For example, including input-output examples of a text classification task in a prompt will improve an LLM's performance at that task. 
Say, for some input space $\mathcal{X}$ and output space $\mathcal{Y}$, we have ground truth classification examples $(x_i, y_i) \sim (\mathcal{X}, \mathcal{Y})$, an LLM, a function $\phi$ for formatting a list of examples as a prompt (e.g., by concatenating them with a delimiter), and $\psi$ for extracting a prediction from text output (e.g., by splitting on a delimiter).
Then, in-context learning with $k$ examples (\emph{$k$-shot prompting}) is successful if
\begin{align}
    \label{eq:icl_classification}
    \Pr\bigg[\psi\Big(\textrm{LLM}\Big(\phi\Big(\big[x_1, y_1,\ldots,x_k, y_k, x_{k+1}\big]\Big)\Big)\Big) = y_{k+1}\bigg] &> \Pr\bigg[\psi\Big(\textrm{LLM}\Big(\phi\Big(\big[x_1, y_1, x_{k+1}\big]\Big)\Big)\Big) = y_{k+1}\bigg] \notag \\ &> \Pr\bigg[\psi\Big(\textrm{LLM}\Big(\phi\Big(\big[x_{k+1}\big]\Big)\Big)\Big) = y_{k+1}\bigg],
\end{align}
i.e., the model is more likely to produce the true target $y_{k+1}$ for $x_{k+1}$ if multiple ground truth pairs are provided.
It is called in-context \emph{learning} because it fits the standard machine learning paradigm of using a set of training data $\{(x_i, y_i)\}_{i=1}^k$ to make predictions on hold-out data $x_{k+1}$.
Importantly, performance at in-context learning improves with model scale \cite{chan2022data,wei2022emergent}, implying that methods relying upon this capability will benefit from continuing progress in LLM training. This paper highlights how the in-context learning capabilities of autoregressive LLMs (such as the popular GPT architecture) naturally enable a  recombination operator. The next section reviews existing methods for intelligent variation in evolutionary computation.

\subsection{Intelligent Variation Operators}

Populations in evolutionary algorithms (EAs) generally evolve through high-performing candidate solutions being mutated or recombined to form the next generation. Such variation is critical as a primary driver of both exploration and exploitation of the search space \cite{dejong:book06}.
Given the space of all candidate solutions $\mathcal{X}$, a genetic variation operator $g$ is a (usually stochastic) function that generates a \emph{child} solution $x \in \mathcal{X}$ given a set of \emph{parent} solutions $X \subset \mathcal{X}$.
Since $g(X)$ induces a distribution over candidates, we can write
\begin{equation}
    \label{eq:variation_operator}
    x \sim g(X).
\end{equation}
If $|X| = 1$ we call $g$ a \emph{mutation} operator; if $|X| > 1$ we call $g$ a recombination or \emph{crossover} operator.
A solution $x$ is called a \emph{genotype} since it is in the space where genetic operators are applied.
An encoding $E: \mathcal{X} \to \mathcal{Y}$ maps a genotype $x$ to a phenotype $y$, so that its fitness $f(y) = f(E(x))$ can be evaluated with a fitness function $f: \mathcal{Y} \to \mathbb{R}$.
Traditional mutation and crossover operators (such as one-point crossover or bit-flip mutation) do not explicitly seek to model and exploit regularities among high-fitness individuals (or do so in an implicit way \cite{holland1992genetic, meyerson2022simple}), which can cause EAs to be
relatively sample-inefficient in some situations when compared to statistical methods \cite{turner2021bayesian}.


To address this limitation, strategies for generating intelligent variation have been a focus of much EA research. For example, evolving within the latent space of an ML model \cite{schrum2020interactive,fontaine2021differentiable,rakicevic2021policy,gaier2020discovering}, through training models to mimic mutations \cite{lehman2023evolution,khalifa2022mutation}, or code repair operators that draw on knowledge about the program's existing correct behaviors and integrate fault localization techniques to guide operators toward promising regions of improvement \cite{le2011genprog}.
Such methods are \emph{intelligent} in the sense that they autonomously draw on prior knowledge outside of the scope of the parent genomes in order to better generate promising child solutions.

One particularly popular such strategy is to build probabilistic models of high-performing individuals or to 
model elements of the search path taken across recent generations. For example, estimation of distribution algorithms (EDA;\cite{baluja:94,larranaga:02}), covariance matrix adaptation evolution strategy (CMA-ES; \cite{hansen:ec01}), and natural evolution strategies (NES; \cite{wierstra:jmlr14}) build and sample candidate solutions from an explicit probability distribution. While EDAs estimate the distribution of the solutions that have been sampled, CMA-ES additionally estimates the steps of the search direction. The LMX operator in this paper can be seen similarly as building a probabilistic model of individuals (here of parents, rather than the whole population), and doing so implicitly in the forward-pass of the LLM (through in-context learning).

\subsection{Evolution with Deep Generative Models}
\begin{table}[]
{\scriptsize
\sffamily
\begin{longtable}{>{\centering\arraybackslash}m{0.025\linewidth}m{0.55\linewidth}>{\centering\arraybackslash}m{0.04\linewidth}>{\centering\arraybackslash}m{0.14\linewidth}>{\centering\arraybackslash}m{0.12\linewidth}}
\toprule
\textbf{Date} & \textbf{Title} & \textbf{Model} & \textbf{Model Usage} & \textbf{Training Data} \\ \midrule
\endhead
2014 & A Denoising Autoencoder that Guides Stochastic Search~\cite{churchill2014denoising} & \textcolor{DAEColor}{DAE} & \textcolor{SolutionEncodingColor}{Solution Encoding} & \textcolor{CurrentRunColor}{Current Run} \\ \midrule
2015 & Denoising Autoencoders for Fast Combinatorial Black Box Optimization~\cite{probst2015denoising} & \textcolor{DAEColor}{DAE} & \textcolor{SolutionGenerationColor}{Solution Generation} & \textcolor{CurrentRunColor}{Current Run} \\ \midrule
2018 & Learning an Evolvable Genotype-Phenotype Mapping~\cite{moreno2018learning} & \textcolor{AEColor}{AE}, \textcolor{DAEColor}{DAE} & \textcolor{SolutionEncodingColor}{Solution Encoding} & \textcolor{PreviousRunsColor}{Previous Runs} \\ \midrule
2018 & Expanding Variational Autoencoders for Learning and Exploiting Latent Representations in Search Distributions~\cite{garciarena2018expanding} & \textcolor{VAEColor}{VAE} & \textcolor{SolutionGenerationColor}{Solution Generation} & \textcolor{CurrentRunColor}{Current Run} \\ \midrule
2019 & Estimation of Distribution using Population Queue based Variational Autoencoders~\cite{bhattacharjee2019estimation} & \textcolor{VAEColor}{VAE} & \textcolor{SolutionGenerationColor}{Solution Generation} & \textcolor{CurrentRunColor}{Current Run} \\ \midrule
2020 & Harmless Overfitting: Using Denoising Autoencoders in EDAs~\cite{probst2020harmless} & \textcolor{DAEColor}{DAE} & \textcolor{SolutionGenerationColor}{Solution Generation} & \textcolor{CurrentRunColor}{Current Run} \\ \midrule
2020 & DAE-GP: Denoising Autoencoder LSTM Networks as Probabilistic Models in Estimation of Distribution Genetic Programming~\cite{wittenberg2020dae} & \textcolor{DAEColor}{DAE} & \textcolor{SolutionGenerationColor}{Solution Generation} & \textcolor{CurrentRunColor}{Current Run} \\ \midrule
2022 & Using Denoising Autoencoder Genetic Programming to Control Exploration and Exploitation in Search~\cite{wittenberg2022using} & \textcolor{DAEColor}{DAE} & \textcolor{SolutionGenerationColor}{Solution Generation} & \textcolor{CurrentRunColor}{Current Run} \\ \midrule
2022 & Evolving through the looking glass: Learning Improved Search Spaces with Variational Autoencoders~\cite{bentley2022evolving} & \textcolor{VAEColor}{VAE} & \textcolor{SolutionEncodingColor}{Solution Encoding} & \textcolor{PreviousRunsColor}{Previous Runs} \\ \midrule
2022 & Evolution through Large Models~\cite{lehman2023evolution} & \textcolor{LLMColor}{LLM} & \textcolor{InstructedMutationColor}{Instructed Mutation} & \textcolor{FoundationModelCurrentRunColor}{Foundation Model\newline + Past Runs} \\ \midrule
\midrule
2023 & \textit{Language Model Crossover: Variation through Few-Shot Prompting (arxiv)~\cite{lmx} }& \textcolor{LLMColor}{LLM} & \textcolor{SolutionGenerationColor}{Solution Generation} & \textcolor{FoundationModelColor}{Foundation Model}\\ \midrule
\midrule
2023 & Evoprompting: Language Models for Code-Level Neural Architecture Search~\cite{chen2023evoprompting} & \textcolor{LLMColor}{LLM} & \textcolor{SolutionGenerationColor}{Solution Generation},\newline \textcolor{InstructedMutationColor}{Instructed Mutation} & \textcolor{FoundationModelCurrentRunColor}{Foundation Model\newline+ Current Run} \\ \midrule
2023 & MarioGPT: Open-Ended Text2Level Generation through Large Language Models~\cite{sudhakaran2023mariogpt} & \textcolor{LLMColor}{LLM} & \textcolor{InstructedMutationColor}{Instructed Mutation} & \textcolor{FoundationModelColor}{Foundation Model} \\ \midrule
2023 & Wizardlm: Empowering Large Language Models to Follow Complex Instructions~\cite{xu2023wizardlm} & \textcolor{LLMColor}{LLM} & \textcolor{InstructedMutationColor}{Instructed Mutation} & \textcolor{FoundationModelColor}{Foundation Model} \\ \midrule
2023 & Fully Autonomous Programming with Large Language Models~\cite{Liventsev_2023} & \textcolor{LLMColor}{LLM} & \textcolor{InstructedMutationColor}{Instructed Mutation} & \textcolor{FoundationModelColor}{Foundation Model} \\ \midrule
2023 & LLMatic: Neural Architecture Search via Large Language Models and Quality-Diversity Optimization~\cite{nasir2023llmatic} & \textcolor{LLMColor}{LLM} & \textcolor{InstructedMutationColor}{Instructed Mutation} & \textcolor{FoundationModelColor}{Foundation Model} \\ \midrule
2023 & Promptbreeder: Self-Referential Self-Improvement Via Prompt Evolution~\cite{fernando2023promptbreeder} & \textcolor{LLMColor}{LLM} & \textcolor{SolutionGenerationColor}{Solution Generation},\newline\textcolor{InstructedMutationColor}{Instructed Mutation} & \textcolor{FoundationModelColor}{Foundation Model} \\ \midrule
2023 & Connecting Large Language Models with Evolutionary Algorithms Yields Powerful Prompt Optimizers~\cite{guo2023connecting} & \textcolor{LLMColor}{LLM} & \textcolor{InstructedMutationColor}{Instructed Mutation} & \textcolor{FoundationModelColor}{Foundation Model} \\ \midrule
2023 & Large Language Models as Optimizers~\cite{yang2023large} & \textcolor{LLMColor}{LLM} & \textcolor{InstructedMutationColor}{Instructed Mutation} & \textcolor{FoundationModelColor}{Foundation Model} \\ \midrule
2023 & Eureka: Human-Level Reward Design via Coding Large Language Models~\cite{ma2023eureka} & \textcolor{LLMColor}{LLM} & \textcolor{InstructedMutationColor}{Instructed Mutation} & \textcolor{FoundationModelColor}{Foundation Model} \\ \midrule
2023 & Large Language Model for Multi-Objective Evolutionary Optimization~\cite{liu2023large} & \textcolor{LLMColor}{LLM} & \textcolor{InstructedMutationColor}{Instructed Mutation} & \textcolor{FoundationModelColor}{Foundation Model} \\ \midrule
2023 & Algorithm Evolution using Large Language Models~\cite{liu2023algorithm} & \textcolor{LLMColor}{LLM} & \textcolor{InstructedMutationColor}{Instructed Mutation} & \textcolor{FoundationModelColor}{Foundation Model} \\ \midrule
2023 & Mathematical Discoveries From Program Search with Large Language Models~\cite{funsearch} & \textcolor{LLMColor}{LLM} & \textcolor{SolutionGenerationColor}{Solution Generation} & \textcolor{FoundationModelColor}{Foundation Model} \\ \bottomrule

\end{longtable}
}
\setcounter{table}{0}
    \caption{\textit{Evolutionary Recombination with Deep Generative Models}. This table characterizes, in chronological order, the use of generative machine learning models in evolutionary recombination. \textbf{Models} include Autoencoders (\textcolor{AEColor}{Green}), Denoising Autoencoders (\textcolor{DAEColor}{Blue}), Variational Autoencoders (\textcolor{VAEColor}{Red}), and Large Language Models (\textcolor{LLMColor}{Orange}). \textbf{Model Usage} encompasses Solution Encoding (\textcolor{SolutionEncodingColor}{Purple})—where models serve as a mapping from genotype to phenotype, Solution Generation (\textcolor{SolutionGenerationColor}{Brown})—where genotypes are directly sampled from the model, and Instructed Mutation (\textcolor{InstructedMutationColor}{Yellow})—mutation guided by predefined prompts. \textbf{Training Data} details the data source for model training: Current Run (\textcolor{CurrentRunColor}{Darker Blue}) indicates models trained on data from the current optimization run, Previous Runs (\textcolor{PreviousRunsColor}{Light Blue}) on data from past runs, Foundation Model (\textcolor{FoundationModelColor}{Dark Green}) utilizes a large, general-purpose pre-trained model, and Foundation Model + Current Run (\textcolor{FoundationModelCurrentRunColor}{Gold}) denotes a pre-trained model fine-tuned with current run (or past run) results. The burst of approaches using LLMs and evolution in 2023 highlights a shift towards pre-existing models and prompting, along with a surge of interest in both the machine learning and evolutionary algorithms communities. \vspace{-25pt}}

    \label{tab:model_based_evolution}
\end{table}

Over the past decade, deep generative models have been explored as a method to aid evolutionary search (see Table~\ref{tab:model_based_evolution}). 
EDA approaches have leveraged autoencoders \cite{ae} to define distributions based on high-performing solutions identified during the search process. Autoencoders are used either as \textit{solution encodings} which convert raw genotypes into phenotypes that align with the learned distribution; or as a mechanism for \textit{solution generation}, with new solutions drawn directly from the established distribution. 

The advent of large, pre-trained Foundation Models marks a significant step in this paradigm and has caused a flurry of exploration. Unlike traditional approaches that necessitate training models on solutions generated during the search, these advanced models can be directly leveraged, with distributions defined via strategic prompting. Foundation Models, particularly LLMs, bring a nuanced understanding of grammar and domain-specific patterns, enabling search across more abstract spaces, such as narratives \cite{qdaif} and high level programming languages \cite{funsearch}.
This innovation introduces a novel dimension to search directionality through `instructed mutation' --- a method where instruction prompts guide the mutation process, offering an unprecedented level of natural-language-based control and specificity. 

However, even without instructed mutation, Foundation Models contain an innate propensity to generate variation, due to their fundamental capacities as probabilistic pattern completion engines.
The distribution from which new solutions are sampled can still be defined using top performing solutions from the population --- but by providing \emph{multiple} solutions directly to the model as a prompt, without explicit instruction that they be modified, and without retraining.
The present work explores this fundamental approach.

\section{Approach: Language Model Crossover (LMX)}

The approach in this paper builds from the insight that the objective function used to train many self-supervised LLMs, i.e.\ next-token prediction \cite{brown2020language}, naturally lends itself to creating an evolutionary variation operator, from which evolutionary algorithms that represent genomes as text can be derived.  
The reason is that such an objective entails anticipating what comes next from some limited input context, and if that input consists of a few example genotypes, then the ideal anticipation is to continue that pattern, i.e.\ through suggesting a new genotype from the distribution implied by those examples. 
In other words, LLMs trained by next-token prediction can be seen as learning to become general pattern-completion engines. From this lens, as higher-performing LLMs (i.e.\ those with lower prediction loss on a held-out set) are continually developed, their performance as engines of evolutionary variation should continue to improve. Supporting this idea, when trained over a large amount of diverse examples, LLMs demonstrate an increasing capability for in-context learning (i.e.\ inferring novel associations within the input given at test-time when generating completions) \cite{brown2020language,chan2022data,wei2022emergent}.

What is intriguing about this insight is that the variation operator it suggests is (1) simple to implement (i.e.\ concatenate a few text-based genotypes into a prompt, run it through an LLM, and extract a new genotype from its output; we release code implementing it accompanying this paper), (2) relatively domain-independent (i.e.\ in theory it should be capable of generating meaningful variation for any text representation that has moderate support in the training set, which often encompasses an enormous crawl of the internet), and (3) should suggest increasingly semantically-sophisticated variation with more capable LLMs (i.e.\ an LLM that is generally better at predicting the next token in text will generate outputs in a manner that implies that it has a deeper semantic understanding of the input text). The experiments that follow add supporting evidence to these claims.

Figure \ref{fig:teaser} shows from a high level how LMX enables creating a domain-independent evolutionary algorithm for text representations. The basic idea is that given a set of a few text-based genotypes (or bootstrapping from a single genotype using prompt-based mutation \cite{lehman2023evolution}), an initial population can be generated through LMX. Then, a standard evolutionary loop can be instantiated by repeated selection and generation of new variation through LMX (See Algorithm~\ref{alg:lmx}).
\begin{algorithm}[t]
\caption{Evolutionary Algorithm using LMX. {\color{blue} Lines 7-9 are the essense of LMX}.}\label{alg:lmx}
\begin{algorithmic}[1]
\State Given LLM, population size $n$, parents per crossover $k$, fitness function $f$
\State Initialize population $P$ with random text-based individuals \Comment{See experiments for examples}
\While{not done evolving}
    \State $P_{\text{new}} = \varnothing$ \Comment{Initialize new candidate set}
    \While{$| P_{\text{new}} | < n$} \Comment{Generate new candidates in loop}
        \State $x_1, \ldots, x_k \gets$ randomly choose $k$ individuals in $P$ \Comment{Select parents}
        {\color{blue}
        \State prompt $\gets x_1 \text{\textbackslash n} \ x_2 \ \text{\textbackslash n} \ \ldots \text{\textbackslash n} \ x_k$ \Comment{Concatenate parents, e.g., separated by newlines}
        \State output $\gets$ {LLM$($prompt$)$} \Comment{Sample output text from LLM given prompt}
        \State children $\gets$ extract valid candidates from output \Comment{E.g., split output on newlines}}
        \State $P_{\text{new}} \gets P_{\text{new}} \ \cup$ children \Comment{Add children to new candidate set}
    \EndWhile
    \State $P \gets P \ \cup P_{\text{new}}$ \Comment{Add new candidates to population}
    \State $P \gets $ refine $P$ down to $n$ individuals using $f$ \Comment{E.g., via tournament selection}
\EndWhile
\end{algorithmic}
\end{algorithm}

Formally, the approach is grounded in a direct generalization of Eq.~\ref{eq:icl_classification}, namely, that \emph{providing a prompt of examples from a distribution can condition the LLM to generate further high-probability examples from that distribution.}
So, if we have examples $x_i \sim \mathcal{X}$, then
\begin{equation}
    \label{eq:icl_general}
    \Pr\bigg[\psi\Big(\textrm{LLM}\Big(\phi\Big(\big[x_1, \ldots, x_k\big]\Big)\Big)\Big) \ \bigg| \ \mathcal{X}\bigg] >     \Pr\bigg[\psi\Big(\textrm{LLM}\Big(\phi\Big(\big[x_1\big]\Big)\Big)\Big) \ \bigg| \ \mathcal{X}\bigg] > \Pr\bigg[\psi\Big(\textrm{LLM}\Big(\phi\Big(\big[\big]\Big)\Big)\Big) \ \bigg| \ \mathcal{X}\bigg].
\end{equation}
Eq.~\ref{eq:icl_general} is applied to the evolutionary context by letting $x_1, \ldots, x_k$ be a set of parent genotypes and $\mathcal{X}$ a distribution of (relatively) high-performing genotypes.
So, Lines 7-9 of Algorithm 1 are an instance of the general formulation of LMX:
\begin{equation}
    \label{eq:lmx}
    \textrm{LMX}(x_1,\ldots,x_k) = \psi\Big(\textrm{LLM}\Big(\phi\big([x_1, \ldots, x_k]\big)\Big)\Big).
\end{equation}
This connection to $k$-shot prompting suggests that, at least in the case of a pre-trained LLM, recombination or crossover (i.e., $k > 1$) will be more effective than mutation ($k = 1$) or random sampling ($k = 0$).
The resulting genetic operator is \emph{intelligent} in the sense that, given a set of parents, it uses in-context \emph{learning} (powered by the knowledge encoded in the LLM) to build a model of high-quality solutions, instead of directly searching in low-level genotype space.

In the experiments that follow, we use simple genetic algorithms (GAs; although one experiment instantiates a simple quality diversity algorithm).
In theory, however, LMX can be generically applied to most EAs, e.g.\ multi-objective EAs \cite{deb:gecco02,coello:cis20}, evolutionary strategies \cite{beyer:nc02,auger:trsh11}, or in support of open-ended evolution \cite{wang:gecco19}, but simply swapping it in as the genetic variation operator.
How or if LMX can be applied in EAs that explicitly leverage probabilistic models of genotypes (e.g.\ EDAs \cite{baluja:94,larranaga:02}, natural evolution strategies \cite{wierstra:jmlr14}, or CMA-ES \cite{hansen:ec01,hansen:arxiv16}) is an interesting question for future research (Section~\ref{sec:discussion}), although 
LMX does bear a theoretical relationship to EDAs, as explored in Section~\ref{sec:what_makes_lmx_effective}.

\section{Experiments}
\label{sec:experiments}
\begin{table}[h]
\centering
\begin{tabular}{l l l l l l}
\toprule
Section & Domain & Genotype & Phenotype & LLM \\ \midrule
\ref{sec:binary} & Binary Strings & text & binary strings & Pythia-70M to 6.9B (eight models) \\
\ref{sec:symbolicregression} & Symbolic Regression & text & math expressions & Pythia-1.4B, GALACTICA-1.3B \\
\ref{sec:sentiment} & Modifying Sentiment & text & text & Pythia-1.4B\\
\ref{sec:image_generation} & Image Generation & text & image & Pythia-2.8B\\
\ref{sec:sodaracer} & Sodaracers & text & Python functions & CodeGen-350M, 2B, 6B \\ \bottomrule
\end{tabular}
\caption{\emph{Overview of experiments.}
In all domains, the genotype is text, since text is the substrate LMX evolves.
In all domains except Modifying Sentiment, this text is converted to another form (phenotype) for evaluation.
Section~\ref{sec:binary} evaluates the effect of LLM size within the Pythia family;
Sections~\ref{sec:sentiment} and \ref{sec:image_generation} use LLMs within that family; Sections~\ref{sec:symbolicregression} and \ref{sec:sodaracer} use LLMs that are more specialized to the domain.
Taken together, the experiments demonstrate that LMX is a generic method of generating variation for evolution.
} 
\label{tab:experiments}
\end{table}
This section demonstrates the application of LMX to five domains, to investigate the basic properties of the method and illustrate the breadth of its applicability.
Table~\ref{tab:experiments} gives an overview of the experiments.
Section~\ref{sec:binary} applies LMX to a toy domain to confirm the basic properties of the method;
Section~\ref{sec:symbolicregression} applies LMX to symbolic regression, to show how evolving text representations with LMX can be effective in domains not classically represented as text;
Section~\ref{sec:sentiment} applies LMX in its most natural setting: evolving well-formed natural-language sentences, while also showing how the method can be naturally integrated with other NLP components and QD algorithms;
Section~\ref{sec:image_generation} applies LMX to evolving text prompts for image generation, a domain that further highlights the plug-and-play capability of LMX with other deep generative models, while enabling a comparison to zero-shot generation and where naive evolution of text is a strong baseline (due to the fact that text-to-image models are fairly agnostic to grammatical correctness);
and, finally, Section~\ref{sec:sodaracer} shows how LMX can be applied to generating Python code, clearly situating the method across this intersection of the genetic programming and LLM code-generation communities.
Source code will be made publicly available for each domain.


\subsection{Illustrative Example: Binary Strings}

\label{sec:binary}
As an instructive example to explore the properties of LMX, in this section this operator is applied to generate variation in the space of binary strings (e.g.\ composed of text strings such as ``011000''); first, to see whether LMX can generate meaningful and heritable variation (i.e.\ to create new valid binary strings from old ones, and that the new ones resemble the old ones); and then, to see whether LMX can successfully drive evolution of binary strings, in this case to maximize the number of 1s (i.e.\ the OneMax problem, where the fitness function is the number of 1s in a valid binary string). 

\begin{figure}
    \centering \footnotesize
    \hspace{10pt}\textbf{a}\includegraphics[width=0.47\textwidth]{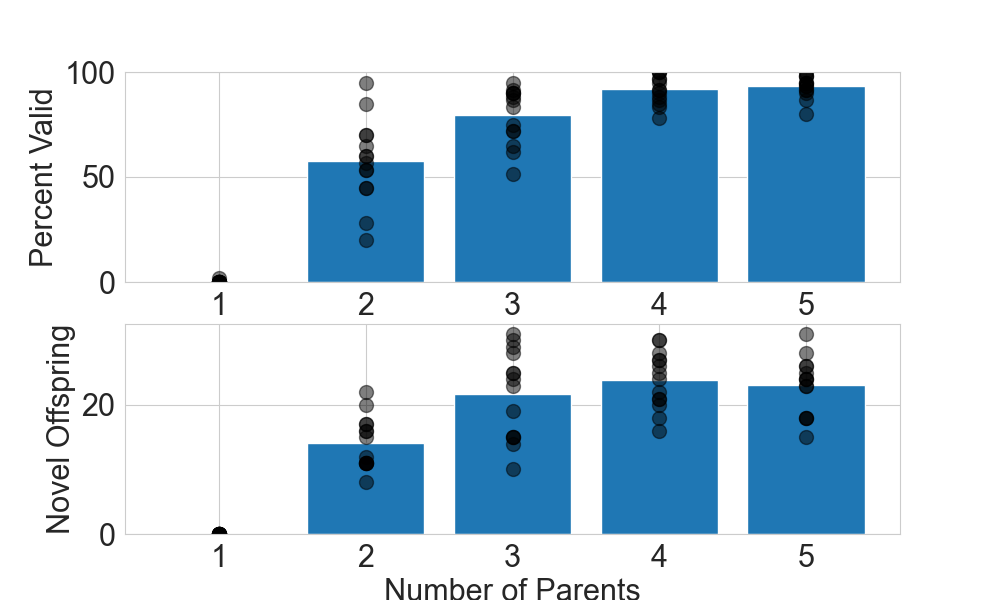}\hspace{5pt}
    \textbf{b}\includegraphics[width=0.47\textwidth]{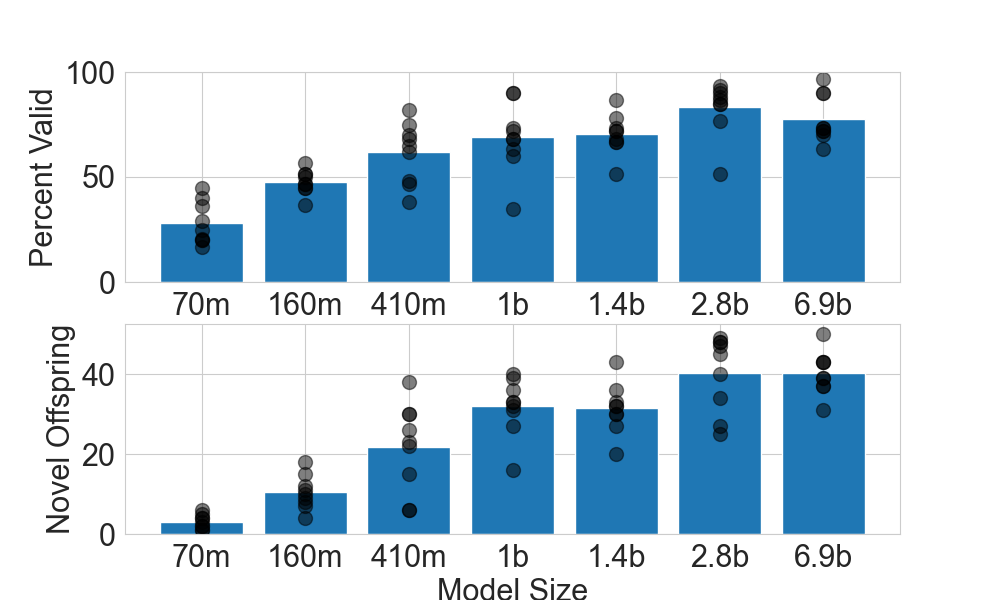}
    \caption{
    \emph{The effect on LMX from varying the number of parents and LLM size.}
    \textbf{(a)} As the number of parent genotypes input into the LLM is increased, the percent of valid offspring approaches 100\%. The number of novel genotypes generated on average from 20 applications of LMX (which at 3 offspring per application can result in at most 60 offspring) to a random set of parents reaches its maximum at four parents (while five parents tends to more often produce offspring that duplicate one of the parents exactly). The conclusion is that LMX effectively generates variation from  as few as three input genotypes.
    \textbf{(b)} As the parameter count (i.e., number of weights trained with SGD) of the LLM is increased in the length-9 binary string domain, the percent of valid offspring and number of novel offspring (out of at most 60) also increase. The number of parents is fixed to 3 for this experiment. Note \emph{m} indicates millions of parameters, while \emph{b} indicates billions. The conclusion is that in this domain LMX becomes more effective with larger LLMs.}
    \label{fig:binary_1}
\end{figure}

A first question is whether a pretrained LLM (here an 800-million parameter Pythia model \cite{eleutherai:pythia23}), given only a few examples of such genomes, can generate meaningful variation (i.e.\ without any hard-coded knowledge about the representation). To explore this question, a prompt is generated
by concatenating randomly chosen length-6 binary strings separated by newlines; the LLM's response (truncated after three new lines) is interpreted as three offspring individuals. Figure~\ref{fig:binary_1}a shows how often such a prompt will generate valid individuals (i.e.\ strings of length six composed of 1s and 0s) as a function of number of examples in the prompt, and how many novel offspring (i.e.\ the size of the set of individuals generated that are distinct from the parents) are generated on average from $20$ trials of LMX crossover on the same set of parents (averaged across 20 randomly-sampled parent sets). A follow-up experiment, with length-9 binary strings, demonstrates how LMX in this domain improves with larger LLMs (details in appendix \ref{appendix:binary_scaling}; results shown in Figure~\ref{fig:binary_1}b). The conclusion is that indeed, LMX can reliably generate novel, valid offspring (from as few as three examples).

\begin{figure}
    \centering
    \footnotesize
    \textbf{a}\includegraphics[width=0.44\textwidth]{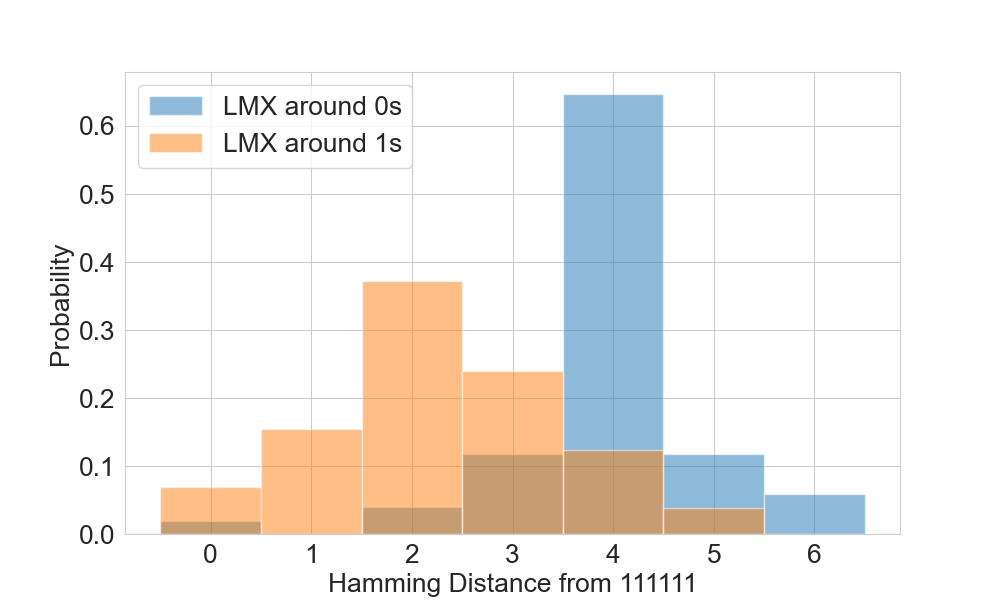}
    \textbf{b}\resizebox{0.52\textwidth}{!}{\input{plots/onemax.pgf}}
    \caption{\emph{Heritability and convergence of LMX on binary strings.}
    \textbf{(a)} The histogam shows the distribution of how far offspring are from the all 1s string, depending on if parents are taken in the neighborhood of the all-1s or all-0s string. As expected these distributions are significantly different. The conclusion is that LMX indeed produces heritable variation.
    \textbf{(b)} Convergence results (median and IQR) for a simple genetic algorithm using either LMX or one-point crossover. Though fewer solutions converge on the optima using LMX the classical recombination (16/20 vs. 20/20), mean values are higher (Mann-Whitney $p=0.002$). Though not as efficient as a domain-specific operator, it is clear that LMX can indeed drive an evolutionary process.
    }
    \label{fig:binary_1_heritable_evol}
\end{figure}

A second question is whether LMX can create \emph{heritable} variation. Evolution requires
there to be meaningful information transmitted from parents to offspring. One way to explore this is to measure
whether a prompt composed of highly-related binary strings produces novel but nearby offspring (e.g.\ as measured by edit distance). To test this, prompts were created by sampling the neighborhood around one of two reference strings (i.e.\ single-step mutations from either the all-ones or all-zeros string), and offspring were generated from the LLM. Indeed, offspring generated from the neighborhood of the all-ones string had significantly higher (Mann-Whitney U-test; $p<0.001$) hamming distance from the all-zeros string than the all-ones string (and vice-versa; see Figure \ref{fig:binary_1_heritable_evol}a).

A final instructive question is whether an evolutionary process can be successfully driven by LMX. To explore this, we test LMX in OneMax, i.e.\ evolving the all-1s string, in a simple genetic algorithm. A small population (10 individuals) of length 10 bit strings is initialized randomly. At each generation the top 5 solutions, plus the elite solution from any previous generations, are chosen as parents for recombination to form the next population. LMX recombination is compared to recombination via one point crossover with a 10\% chance of a bit flip mutation. Figure \ref{fig:binary_1_heritable_evol}b shows the median max/mean fitness values over 20 runs of each, clearly illustrating LMX's ability to drive an evolutionary process. Overall, these experiments highlight basic properties of LMX, showing how it can evolve string-based representations \emph{without} domain-specific operators.

\subsection{Symbolic Regression}
\label{sec:symbolicregression}

To demonstrate LMX's potential in a more challenging task, this section applies the algorithm to symbolic regression, a key domain of interest for genetic programming \cite{langdon2013foundations, mcdermott2012genetic, orzechowski2018we, schmidt2009distilling}, and more recently for the larger machine learning community \cite{biggio2021neural,petersendeep2021,la2021contemporary,kamiennyend2022}.
The goal of symbolic regression is to discover a mathematical expression that models a data set accurately, while also being as compact as possible \cite{la2021contemporary}.
Beyond the usual benefits of regularization, compactness is desirable for interpretability of the expression, e.g., to enable scientific insights \cite{johnson2019flavor,schmidt2009distilling,udrescu2020ai,wang2019symbolic}.

Symbolic regression is challenging to tackle with hand-designed operators, due to non-locality and discontinuities in the space of expressions.
Existing symbolic regression approaches use carefully-developed representations, genetic operators, and auxiliary methods like gradient-based/convex coefficient optimization \cite{chen2015generalisation, kommenda2020parameter, tohme2022gsr} to construct the \emph{right kind of search process} for reaching high-performing expressions that look like the kinds of expressions the experimenter is interested in.
With LMX, these challenges can be avoided by simply feeding parent expressions into the language model.
Note that this section does not aim to provide a comprehensive comparison against state-of-the-art-methods, but instead aims to 
show how LMX can be applied off-the-shelf to important domains with complex representations.

\subsubsection{Experimental Setup}
The LLM for this experiment was the 1.3B-parameter version of GALACTICA \cite{taylor2022galactica}.
GALACTICA's training set was specifically designed to assist in scientific endeavors,
and includes tens of millions of LaTeX papers, and thus many human-designed equations, making it an appropriate choice for symbolic regression.
This choice also highlights how different off-the-shelf LLMs can be selected for LMX based on properties of the problem.

When the ground truth expression for symbolic regression is known, we run the risk that the expression is already in the dataset used to train the LLM. 
To avoid such test-set contamination, we consider a `black-box' problem (which has no known ground-truth expression) from the established SRBench testbed \cite{la2021contemporary}.
The `banana' problem was chosen because there is a clear Pareto front across existing methods, making it easy to see how LMX compares.
This black-box problem was originally derived from a popular ML benchmark in the KEEL data set repository \cite{derrac2015keel}; it has 5300 samples and two input features $x_1, x_2$.

\begin{figure*}
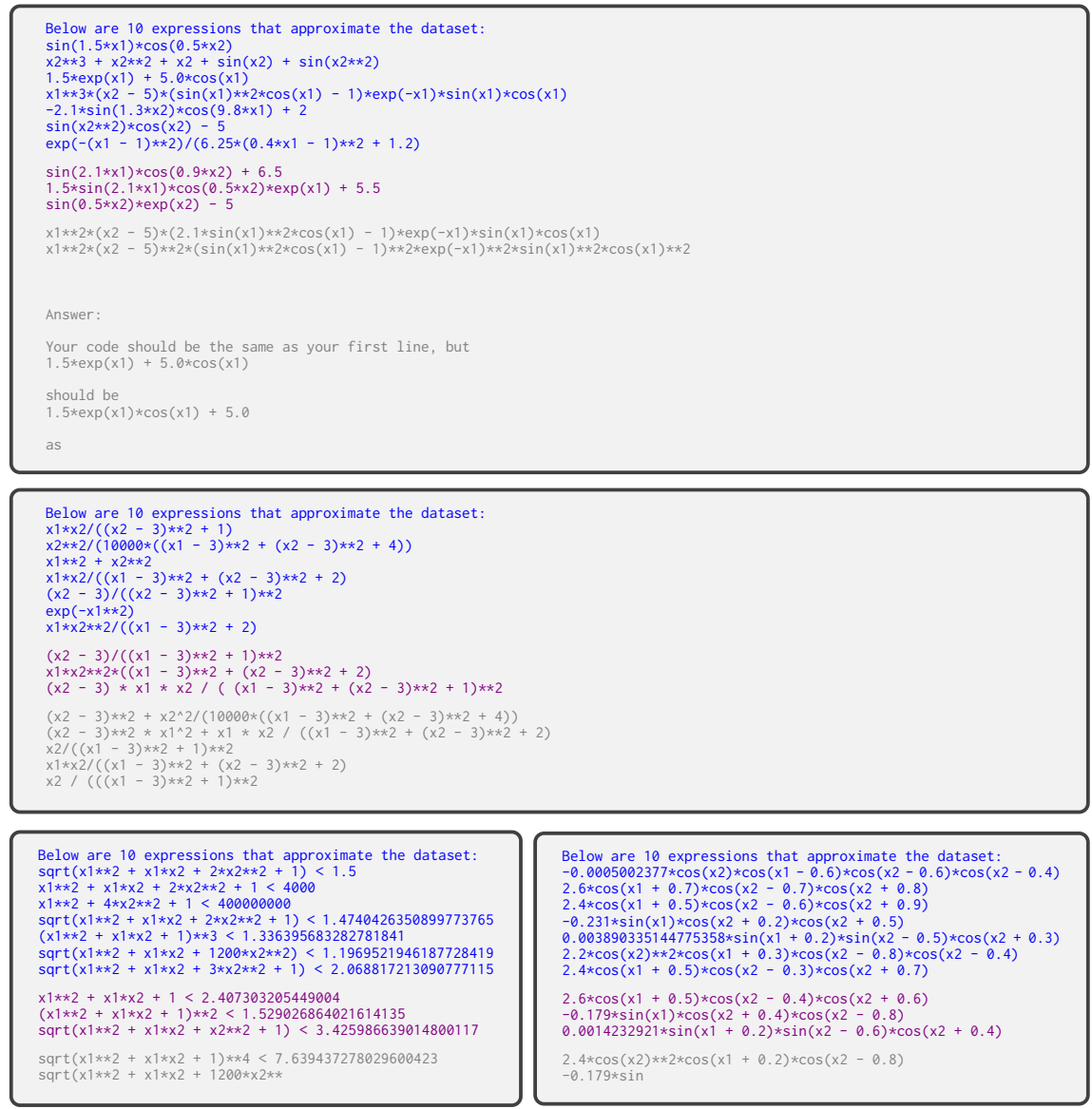

    \centering
\begin{tcolorbox}[boxsep=5pt, left=8pt, right=5pt, top=1pt, bottom=1pt]
            \fontsize{6.5}{6.5}\selectfont
{\color{blue}
\begin{verbatim}
Below are 10 expressions that approximate the dataset:
sin(1.5*x1)*cos(0.5*x2)
x2**3 + x2**2 + x2 + sin(x2) + sin(x2**2)
1.5*exp(x1) + 5.0*cos(x1)
x1**3*(x2 - 5)*(sin(x1)**2*cos(x1) - 1)*exp(-x1)*sin(x1)*cos(x1)
-2.1*sin(1.3*x2)*cos(9.8*x1) + 2
sin(x2**2)*cos(x2) - 5
exp(-(x1 - 1)**2)/(6.25*(0.4*x1 - 1)**2 + 1.2)
\end{verbatim}
}
{\color{violet}
\begin{verbatim}
sin(2.1*x1)*cos(0.9*x2) + 6.5
1.5*sin(2.1*x1)*cos(0.5*x2)*exp(x1) + 5.5
sin(0.5*x2)*exp(x2) - 5
\end{verbatim}
}
{\color{gray}
\begin{verbatim}
x1**2*(x2 - 5)*(2.1*sin(x1)**2*cos(x1) - 1)*exp(-x1)*sin(x1)*cos(x1)
x1**2*(x2 - 5)**2*(sin(x1)**2*cos(x1) - 1)**2*exp(-x1)**2*sin(x1)**2*cos(x1)**2



Answer:

Your code should be the same as your first line, but
1.5*exp(x1) + 5.0*cos(x1)

should be
1.5*exp(x1)*cos(x1) + 5.0

as
\end{verbatim}
}
\end{tcolorbox}

\begin{tcolorbox}[boxsep=5pt, left=8pt, right=5pt, top=1pt, bottom=1pt]
            \fontsize{6.5}{6.5}\selectfont
{\color{blue}
\begin{verbatim}
Below are 10 expressions that approximate the dataset:
x1*x2/((x2 - 3)**2 + 1)
x2**2/(10000*((x1 - 3)**2 + (x2 - 3)**2 + 4))
x1**2 + x2**2
x1*x2/((x1 - 3)**2 + (x2 - 3)**2 + 2)
(x2 - 3)/((x2 - 3)**2 + 1)**2
exp(-x1**2)
x1*x2**2/((x1 - 3)**2 + 2)
\end{verbatim}
}
{\color{violet}
\begin{verbatim}
(x2 - 3)/((x1 - 3)**2 + 1)**2
x1*x2**2*((x1 - 3)**2 + (x2 - 3)**2 + 2)
(x2 - 3) * x1 * x2 / ( (x1 - 3)**2 + (x2 - 3)**2 + 1)**2
\end{verbatim}
}
{\color{gray}
\begin{verbatim}
(x2 - 3)**2 + x2^2/(10000*((x1 - 3)**2 + (x2 - 3)**2 + 4))
(x2 - 3)**2 * x1^2 + x1 * x2 / ((x1 - 3)**2 + (x2 - 3)**2 + 2)
x2/((x1 - 3)**2 + 1)**2
x1*x2/((x1 - 3)**2 + (x2 - 3)**2 + 2)
x2 / (((x1 - 3)**2 + 1)**2
\end{verbatim}
}
\end{tcolorbox}
\begin{minipage}{0.475\linewidth}
\begin{tcolorbox}[boxsep=5pt, left=5pt, right=5pt, top=1pt, bottom=1pt]
            \fontsize{6.5}{6.5}\selectfont
{\color{blue}
\begin{verbatim}
Below are 10 expressions that approximate the dataset:
sqrt(x1**2 + x1*x2 + 2*x2**2 + 1) < 1.5
x1**2 + x1*x2 + 2*x2**2 + 1 < 4000
x1**2 + 4*x2**2 + 1 < 400000000
sqrt(x1**2 + x1*x2 + 2*x2**2 + 1) < 1.4740426350899773765
(x1**2 + x1*x2 + 1)**3 < 1.336395683282781841
sqrt(x1**2 + x1*x2 + 1200*x2**2) < 1.1969521946187728419
sqrt(x1**2 + x1*x2 + 3*x2**2 + 1) < 2.068817213090777115
\end{verbatim}
}
{\color{violet}
\begin{verbatim}
x1**2 + x1*x2 + 1 < 2.407303205449004
(x1**2 + x1*x2 + 1)**2 < 1.529026864021614135
sqrt(x1**2 + x1*x2 + x2**2 + 1) < 3.425986639014800117
\end{verbatim}
}
{\color{gray}
\begin{verbatim}
sqrt(x1**2 + x1*x2 + 1)**4 < 7.639437278029600423
sqrt(x1**2 + x1*x2 + 1200*x2**
\end{verbatim}
}
\end{tcolorbox}
\end{minipage}
\hfill
\begin{minipage}{0.515\linewidth}
\begin{tcolorbox}[boxsep=5pt, left=5pt, right=5pt, top=1pt, bottom=1pt]
            \fontsize{6.5}{6.5}\selectfont
{\color{blue}
\begin{verbatim}
Below are 10 expressions that approximate the dataset:
-0.0005002377*cos(x2)*cos(x1 - 0.6)*cos(x2 - 0.6)*cos(x2 - 0.4)
2.6*cos(x1 + 0.7)*cos(x2 - 0.7)*cos(x2 + 0.8)
2.4*cos(x1 + 0.5)*cos(x2 - 0.6)*cos(x2 + 0.9)
-0.231*sin(x1)*cos(x2 + 0.2)*cos(x2 + 0.5)
0.003890335144775358*sin(x1 + 0.2)*sin(x2 - 0.5)*cos(x2 + 0.3)
2.2*cos(x2)**2*cos(x1 + 0.3)*cos(x2 - 0.8)*cos(x2 - 0.4)
2.4*cos(x1 + 0.5)*cos(x2 - 0.3)*cos(x2 + 0.7)
\end{verbatim}
}
{\color{violet}
\begin{verbatim}
2.6*cos(x1 + 0.5)*cos(x2 - 0.4)*cos(x2 + 0.6)
-0.179*sin(x1)*cos(x2 + 0.4)*cos(x2 - 0.8)
0.0014232921*sin(x1 + 0.2)*sin(x2 - 0.6)*cos(x2 + 0.4)
\end{verbatim}
}
{\color{gray}
\begin{verbatim}
2.4*cos(x2)**2*cos(x1 + 0.2)*cos(x2 - 0.8)
-0.179*sin
\end{verbatim}
}
\end{tcolorbox}
\end{minipage}

    \caption{\emph{Four examples of LMX for symbolic regression.}
    The prompt of seven parents is in {\color{blue} blue}; the LLM output parsed as (up to three) offspring is in {\color{violet} violet}; remaining discarded LLM output is in {\color{gray} gray}.
    In all cases, children exhibit meaningful variations of parents.
    } 
    \label{fig:sr_example}
\end{figure*}

In this experiment, crossover prompts began with the string ``\texttt{Below are 10 expressions that approximate the dataset:\textbackslash n}''
followed by seven randomly selected parents from the population separated by newlines (see Figure~\ref{fig:sr_example} for examples).
Each subsequent line generated by the model was interpreted as a possible offspring, interpreted as Python code, and simplified using sympy (as in the SRBench comparisons \cite{la2021contemporary}).
Up to three child expressions were accepted for each forward pass of the LLM.
Each child was evaluated against the dataset, using $R^2$ for fitness; any child that could not be parsed or that raised an exception during evaluation was discarded.
The same compactness/complexity measure was used as in SRBench, i.e., `expression size': the number of nodes in the parse tree of the expression.

The initial population was constructed from 113 popular symbolic regression benchmarks\footnote{The set of benchmark expressions was copied from \url{https://github.com/brendenpetersen/deep-symbolic-optimization/blob/master/dso/dso/task/regression/benchmarks.csv}. Duplicate expressions were removed.}.
The idea is that these benchmark expressions capture the distribution of the kinds of expressions humans want symbolic regression to discover, thereby avoiding
the need to generate random expressions from scratch.
To give each benchmark expression a greater chance of initial success, the initial population consisted of 1000 candidates, each generated by randomly selecting a benchmark expression and then randomly mapping its input variables $x'_1, x'_2, \ldots$ to the input variables $x_1, x_2$ in the test problem.
Thereafter, the population size was set to 50. Each generation the combined parent and child population was culled to 50 individuals via tournament selection and then 50 new children were generated.
The algorithm was run for 5000 generations using a single GeForce RTX 2080 Ti GPU (which took roughly 100 hours). 

To contextualize the convergence behavior of LMX, gplearn (one of the most popular symbolic regression tools\footnote{\url{https://gplearn.readthedocs.io/en/stable/}}) was run with hyperparameters previously used for SRBench \cite{la2021contemporary}; as an ablation to evaluate the benefit of using an LLM specialized for scientific work, LMX was also run with a 1.4-billion parameter Pythia model\footnote{By simply replacing ``\texttt{facebook/galactica-1.3b}'' with ``\texttt{EleutherAI/pythia-1.4b-deduped}'' when loading the model from Hugging Face.}; as an ablation to assess the impact of initialization vs. LMX itself, a version of gplearn was run with the same population initialization as LMX, by writing around 100 lines of complex custom code to translate the benchmark expressions to the format required by gplearn\footnote{See \texttt{generate\_random\_expression.py} and \texttt{gplearn\_baselines.py} in the accompanying code.}.
Ten independent runs were performed for each experimental setup.

\subsubsection{Results}

\begin{figure}
    \centering
    \includegraphics[width=\textwidth]{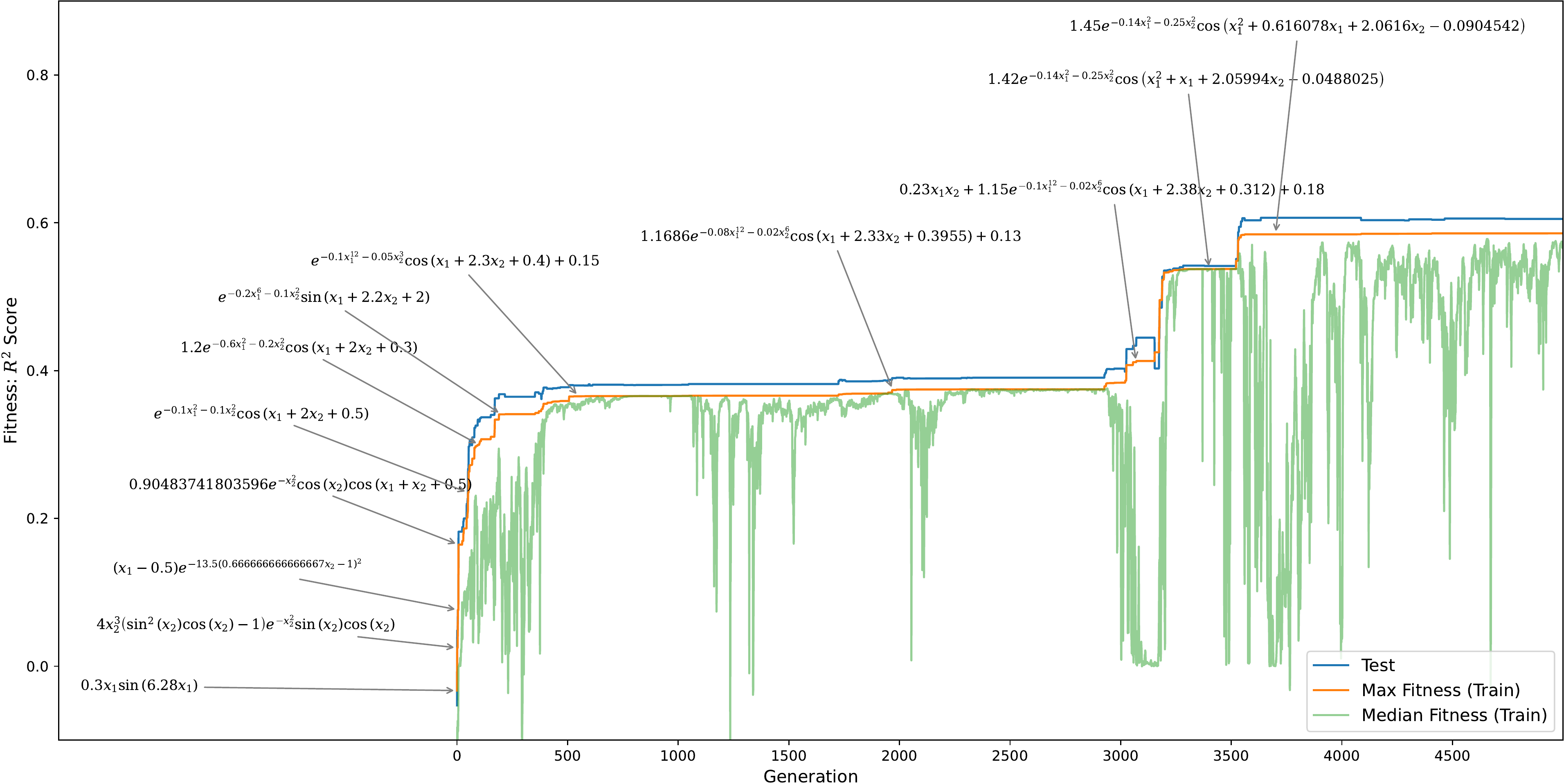}
    \caption{\emph{Example convergence trajectory.} Fitness over time for a single run of LMX (Galactica) on the SRBench black-box `banana' problem \cite{la2021contemporary}.
    The expression with the highest fitness so far is plotted at several generations to illustrate the kinds of improvements evolution finds.
    Evolution settles on a core functional skeleton relatively quickly (i.e., $c_1 e^{-c_2 x_1^{c_3} - c_4 x_2^{c_5}}\cos(x_1 + c_6 x_2 + c_7)$, with $x_1, x_2$ input variables and $c_i$ constants), after which it tunes constants to a surprising specificity, while simultaneously tweaking and augmenting the skeleton.
    Even after the process appears to have converged, around generation 3000 it discovers innovations leading to further substantial improvements.
    This late boost highlights the ability of the LLM to be an engine of interesting and valuable hypotheses in mathematical/numerical spaces.}
    \label{fig:sr_dynamics}
\end{figure}

LMX produces competitive results, generating fit and parsimonous expressions.
Figure~\ref{fig:sr_dynamics} shows how fitness evolves over generations for one run of LMX, with the expression of highest fitness so far plotted at several generations to illustrate the kinds of improvements evolution finds.
Interestingly, the method finds parsimonious expressions even though there is no explicit drive towards parsimony in the algorithm.
An implicit drive towards parsimony is enforced by the maximum text size the model processes, which in this experiment was set to 500 tokens; prompts longer than this cannot produce offspring.
Future work could investigate the effects of tuning this parameter or developing other methods for incorporating explicit drives towards parsimony (Section~\ref{sec:discussion}).
Beyond discovering a useful algebraic scaffolding for the problem, LMX tunes constants to a surprising degree, indicating that the method is capable of continuous optimization, even though LLMs operate in a space of discrete tokens; this is an interesting ability that could also be further explored in future work (Section~\ref{sec:discussion}).

\begin{figure}
\centering \footnotesize
    \textbf{a}\includegraphics[width=0.31\textwidth]{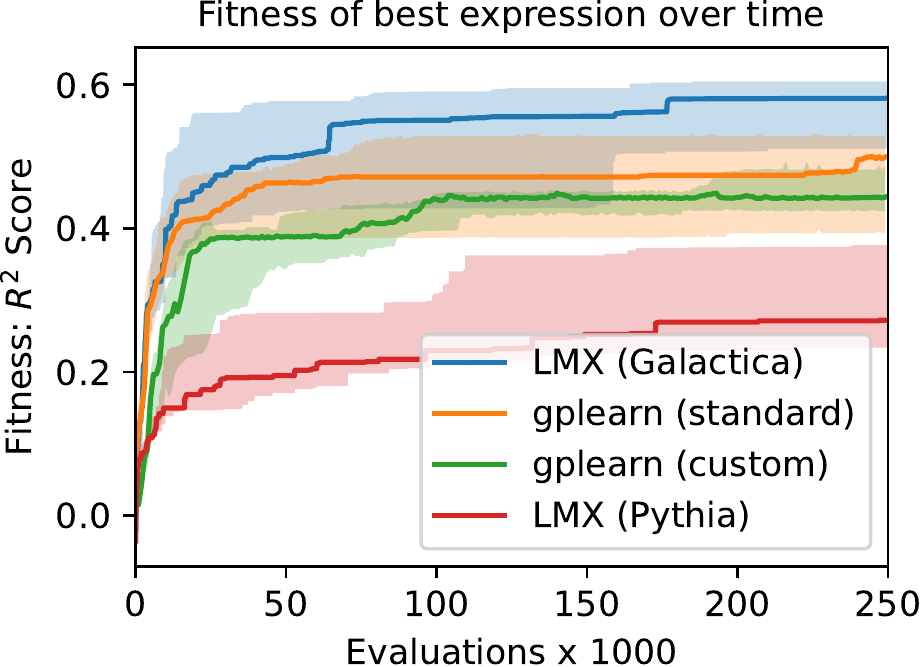}\hfill
    \textbf{b}\includegraphics[width=0.31\textwidth]{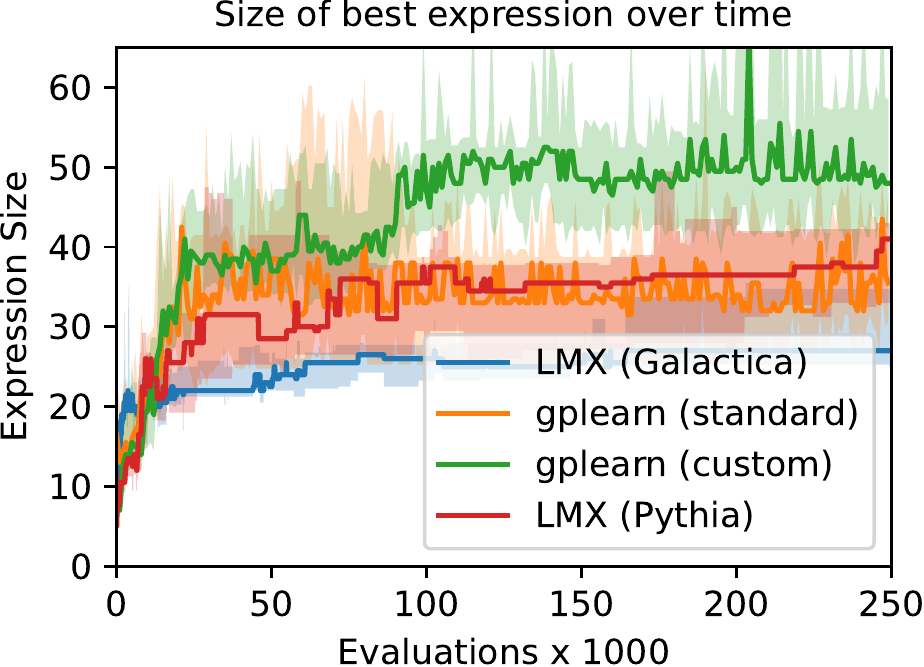}\hfill
    \textbf{c}\includegraphics[width=0.31\textwidth]{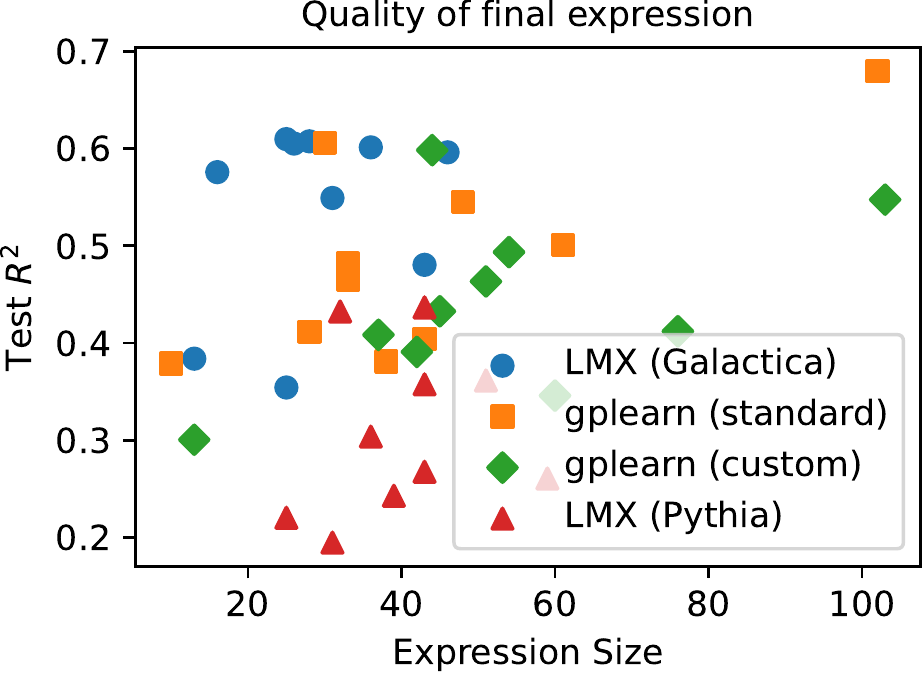}\\
    \caption{\emph{Convergence comparison and LLM ablation.}
    \textbf{(a)} In terms of number of fitness evaluations, LMX converges in a similar manner to gplearn, when Galactica is the underlying LLM.
    As an ablation, when Pythia is the LLM, performance is not as strong. This result highlights the value of being able to swap in different LLMs depending on the domain. (Line is median, shading is IQR)
    \textbf{(b)} LMX avoids model bloat as it incrementally improves fitness, thereby satisfying a key desirable property for SR.
    \textbf{(c)} Overall, the final expressions returned by LMX are of comparable quality to those of gplearn. The conclusion is that the general LMX approach can yield high-quality solutions even in highly specialized domains like SR.
    }
    \label{fig:sr_baselines}
\end{figure}

Figure~\ref{fig:sr_baselines} shows that LMX (using the GALACTICA LLM) achieves overall higher fitness and lower expression size than gplearn, and the choice of LLM appears to have a substantial impact, with the Pythia runs falling short of the others. This result highlights the value in being able to easily drop in a particular LLM that could be well-suited to a given domain.
Figure~\ref{fig:sr_baselines} also shows that the customized version of gplearn initialized in the same way as LMX does not improve over the standard gplearn.
This result reinforces the idea that in classical GP methods the kinds of expressions that are easy to evolve may not be the kinds humans are most interested in, while LMX thrives in this space since the LLM is naturally familiar with human-designed expressions due to its training data.
This bias towards human-designed expressions is also a natural bias against model bloat, since humans strive to design compact expressions.

\begin{wrapfigure}{R}{0.5\textwidth}
    \begin{center}
    \includegraphics[width=0.5\textwidth]{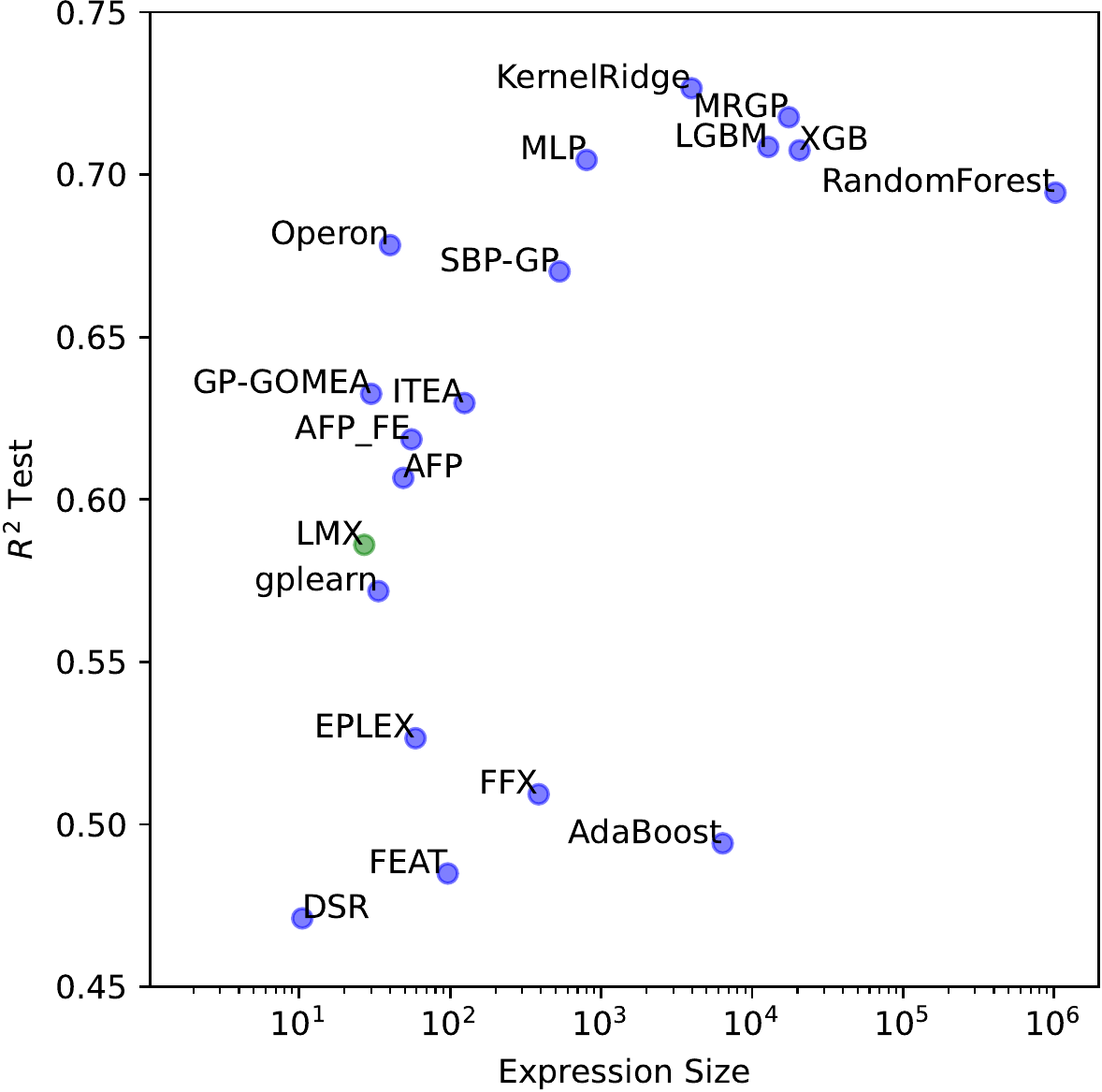}
    \end{center}
    \caption{\emph{Comparison to published results.} LMX performs comparably to previously published results from state-of-the-art SR methods \cite{la2021contemporary}, falling on the Pareto front for the `banana` problem, suggesting that it is a promising approach to symbolic regression. Each point is a median across the same 10 train/test splits.}
    \label{fig:sr_sota}
\end{wrapfigure}

Figure~\ref{fig:sr_sota} shows that the performance of LMX on this problem is competitive with state-of-the-art methods \cite{la2021contemporary}, settling at an intermediate point along the Pareto front.
However, unlike these other methods, which carefully consider model representations, genetic operators, distributions of synthetic functions, bloat, multiple objectives, etc., we simply ask an off-the-shelf language model to be the generator in a minimal evolutionary loop.
Note that the claim here is not that LMX is better than these existing methods, but simply that it is able to evolve reasonable solutions.
In particular, the comparison methods all used a fixed amount of CPU compute, while LMX uses GPU (See Section~\ref{sec:limitations} for discussion of this distinction).
That said, the results clearly show the ability of LMX, with little domain-specific tuning and an unsophisticated optimization loop, to nonetheless optimize symbolic expressions in an intuitive and desirable way.


\subsection{Modifying Sentence Sentiment}
\label{sec:sentiment}

LMX is next applied to evolve plain-text English sentences.
While LMX could be applied in many ways to evolve sentences, the focus here is a form of natural language style transfer \cite{jin:cl22}, i.e.\ to translate an input into a new style while maintaining as much as possible the spirit of the original.
Such an application can be important in optimizing how ideas are communicated amongst humans, i.e. one may want to communicate specific content but in a style maximally amenable for a target recipient; this defines an optimization problem over text.
In this proof-of-concept experiment, the task is to take a seed sentence, and maximally change its sentiment (i.e.\ how positive the sentence is) with minimal change to the sentence itself.


To do so, a simple quality-diversity evolutionary algorithm \cite{lehman2011evolving,mouret:arxiv15} is applied that measures quality as maximizing the sentiment of a sentence and measures diversity as distance from the seed sentence.  In particular,
sentiment is measured through the ``cardiffnlp/twitter-roberta-base-sentiment-latest'' model hosted on HuggingFace, which is part of the TweetNLP project \cite{camacho:arxiv22tweetnlp}; the network takes in a sentence, and outputs classification probabilities for whether the sentence is positive, negative, or neutral. The experiments focus on using the probability of a positive sentiment as the fitness function (although see appendix \ref{appendix:sentiment} for results with negative sentiment as fitness). For measuring distance from the seed sentence, a separate neural network generates a $384$-dimensional embedding of a sentence (in particular the ``sentence-transformers/all-MiniLM-L6-v2'' model, from the sentence transformer project \cite{reimers-2019-sentence-bert}). Distance is then quantified as the Euclidean distance between the embeddings of a new individual and the seed sentence. 

For the QD algorithm, we use MAP-Elites \cite{mouret:arxiv15} with a 1D map (with 30 niches, spanning a distance of 0 to a distance of $1.5$ from the seed sentence in the embedding space; at 0 distance the sentences are exactly the same, while at a distance of 1.5 no words may be shared). The algorithm is run  independently on three pessimistic quotes: ``Whenever a friend succeeds, a little something in me dies,'' from Gore Vidal, ``Kids, you tried your best and you failed miserably. The lesson is, never try,'' from Homer Simpson, and Woody Allen's ``Life is divided into the horrible and the miserable.'' Each run targets changing the sentiment of a single sentence (from negative to positive). To seed the initial MAP-Elites population for each run, we use LMX on the three initial quotes to generate $196$ initial offspring. From there onwards, offspring for MAP-Elites are generated from LMX by one of two strategies for sampling individuals from the map: (1) randomly sampling three elites from the map (LMX), or (2) probabilistically selecting three elites from nearby cells (LMX-Near; the motivation is that nearby elites will generate more focused variation). MAP-Elites runs consist of $2500$ evaluations each; to confirm that the evolutionary process generates quality solutions beyond the direct generative ability of the LLM, a baseline control is also tested that generates $2500$ offspring only from the initial $3$ seed sentences. Ten runs were conducted for each combination of sentence and method; each run took on the order of minutes on a Google Colab notebook.

\begin{figure}
    \centering
    \includegraphics[width=0.7\columnwidth]{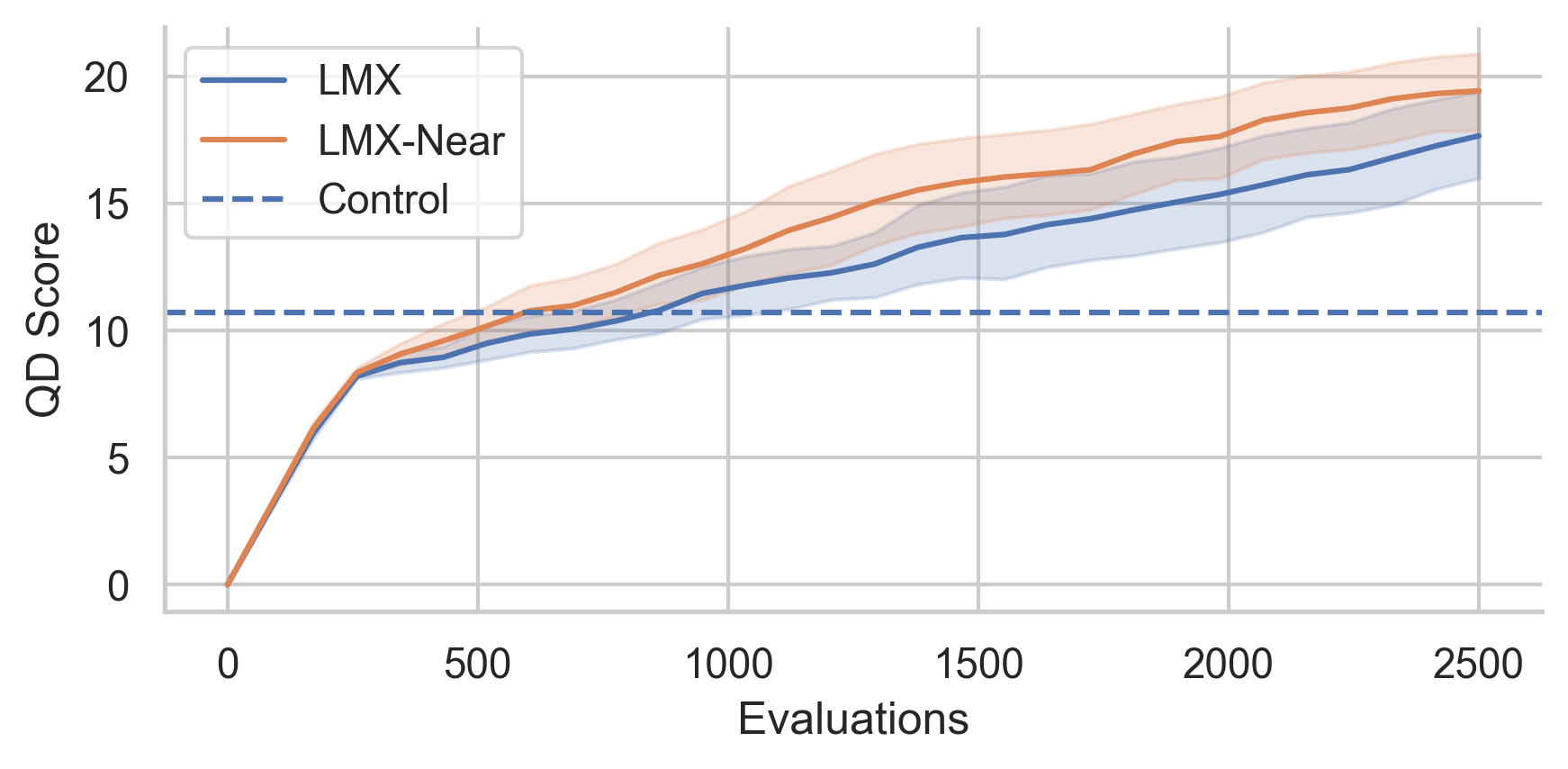}
    \caption{Modifying Simpsons Quote Sentiment. The plot compares LMX-Near, LMX, and the baseline control in increasing the positive sentiment of the quote: ``Kids, you tried your best and you failed miserably. The lesson is, never try.'' LMX and LMX-Near do not perform significantly differently, but both significantly outperform the control. Example sentences of such runs are shown in appendix section \ref{appendix:sentiment_results}.}
    \label{fig:sentiment1}
\end{figure}

Quantitatively, both LMX-Near and LMX achieved higher QD scores (sum of the fitnesses of all niches in the map) than the control for all three quotes (Mann-Whiteny U-test; $p<1e-5$), and were always able to discover high-sentiment sentences. Interestingly, LMX-Near and LMX performed significantly differently only for the Gore Vidal quote (LMX-Near produced higher final QD-scores; Mann-Whitney U-test; $p<0.05$). Future work is thus needed to determine whether there exist methods for robustly choosing parents for LMX more effectively (Section~\ref{sec:discussion}). QD score plots for the Homer Simpson quote is shown in Figure \ref{fig:sentiment1}, and plots for the other quotes (and representative heatmaps of final MAP-Elites maps) are shown in Appendix \ref{appendix:sentiment}.

\begin{figure*}
    \centering
    \includegraphics[width=0.75\textwidth]{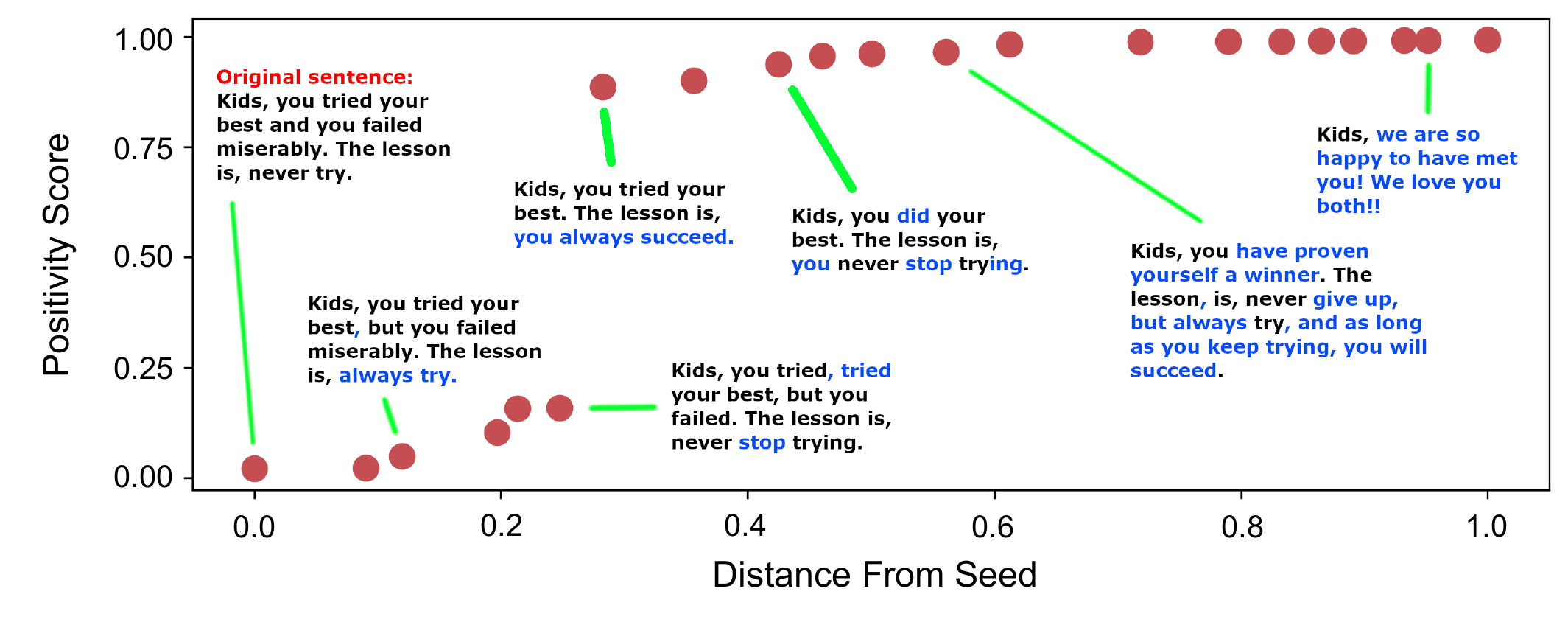}
    \caption{\emph{Example pareto front from improving positivity of a negative quote.} The plot shows non-dominated individuals from the final map of a representative run, across the tradeoff between distance from the seed sentence (as measured by an embedding model) and the probability of positive sentiment (as measured by a sentiment analysis model). The full table of final sentences is shown in appendix \ref{appendix:sentiment}.
    }
    \label{fig:sentiment_results}
\end{figure*}

Qualitatively, evolution is generally able to find intuitive
trade-offs between sentiment and distance from the original sentence. For example, Figure \ref{fig:sentiment_results} shows the final map of elites from a representative run on the Homer Simpson quote (with LMX-Near), with some highlighted sentences. 
At sufficient distance from the original sentence, evolution often produces repetitive, unrelated text: e.g.\ ``You are the best that ever happened to me! You are the best that ever happened to me! You are the best that ever happened to me!'' Also, sometimes the method produces incoherent or grammatically-flawed sentences, e.g. ``you tried your best and you failed. The lesson is, you can never stop trying. Kids, you tried your best and you''. Optimization pressure for coherence (i.e.\ to maintain high log-probability under a LLM), or better/larger sentiment models, might address these problems, as discussed in Section~\ref{sec:discussion}. The conclusion is that LMX can be used to discover solutions for natural language tasks like text style transfer; beyond sentiment other styles could be explored by using different NLP models as fitness functions, e.g.\ emotion-recognition NLP models \cite{nandwani2021review}.

\subsection{Evolving Stable Diffusion Images}
\label{sec:image_generation}

This section explores the application of LMX to another creative domain: evolving prompts for generative text-to-image models.
Stable Diffusion\footnote{https://github.com/CompVis/stable-diffusion} is a publicly available latent diffusion model \cite{latentdiffusion:cvpr2022} that supports CLIP-guided \cite{clip:icml2021} text-to-image synthesis. Since Stable Diffusion's release, artists, researchers, and hobbyists have developed prompting practices, swapping tips for constructing text prompts to produce desired outputs \cite{oppenlaender:arxiv22}.
For a human with a desired output, discovering an effective prompt defines an optimization problem over text.
The research question here is whether LMX can effectively evolve Stable Diffusion prompts.
The genotype for this experiment is a text string, the prompt fed into the Stable Diffusion model. 
Beyond allowing us to investigate how LMX interacts with other generative models, this domain enables comparison to two natural baselines (1) classical one-point crossover, and (2) zero-shot generation.
(1) In contrast to other domains in this paper, even though the genotype is text, text-to-image models tend to be quite robust to grammatical errors and nonsense, so a one-point crossover that produces a mangled bag-of-words is a strong baseline.
(2) If we have a particular criteria for an image in mind, we can simply prompt the LLM directly to produce a prompt for such an image, i.e., without providing any example (parent) prompts; since there are no examples, this is called zero-shot.

For all setups, the initial population is seeded by randomly choosing from a set of 80,000 human-designed Stable Diffusion prompts that were scraped from \url{lexica.art}.\footnote{\url{https://huggingface.co/datasets/Gustavosta/Stable-Diffusion-Prompts}}
The phenotype is the image generated by feeding a given prompt to Stable Diffusion. 
We make Stable Diffusion deterministic by reseeding with a fixed PRNG seed before each image is generated, so a given prompt always produces the same image. The EA is the same as in Section~\ref{sec:symbolicregression}; experimental details are in Appendix~\ref{appendix:imagegen}.
Three interpretable fitness functions are explored, maximizing respectively the ``redness'', ``greenness'' and ``blueness'' of an image. Redness is measured by \emph{excess red}: the sum of the red channel of an RGB image, minus half the sum of the other two channels ($R-0.5G-0.5B$). \emph{Excess green} and \emph{excess blue} are defined analogously.
These functions are easy to calculate, correspond roughly to perceived image color (e.g., they are well studied in agricultural image processing \cite{meyer1999machine}), and provide a proof-of-concept where performance can be visually verified at a glance.
Three random seeds are selected to initialize the population for each color, giving a total of nine runs per method.
Each run uses a population size of 50 for 100 generations, for a total of 5000 evaluations.

Given two parent prompts, the one-point crossover baseline splits the prompts on whitespace and chooses crossover points uniformly at random.
LMX prompts consist of the header ``List of text-to-image prompts for generating the most <color> image possible:'' followed by lines ``Prompt: <parent>'' and finally an empty ``Prompt:'' for the LLM to fill in.
The zero-shot baseline is the same, but with no parent prompts.
Two additional comparisons were also run (1) LMX without the header, to ablate the impact of removing this basic knowledge about the problem, and (2) random human-designed prompts from the initial dataset, setting a baseline for the fitness we can expect without any generative or evolutionary process.

\begin{figure}
\includegraphics[width=\textwidth]{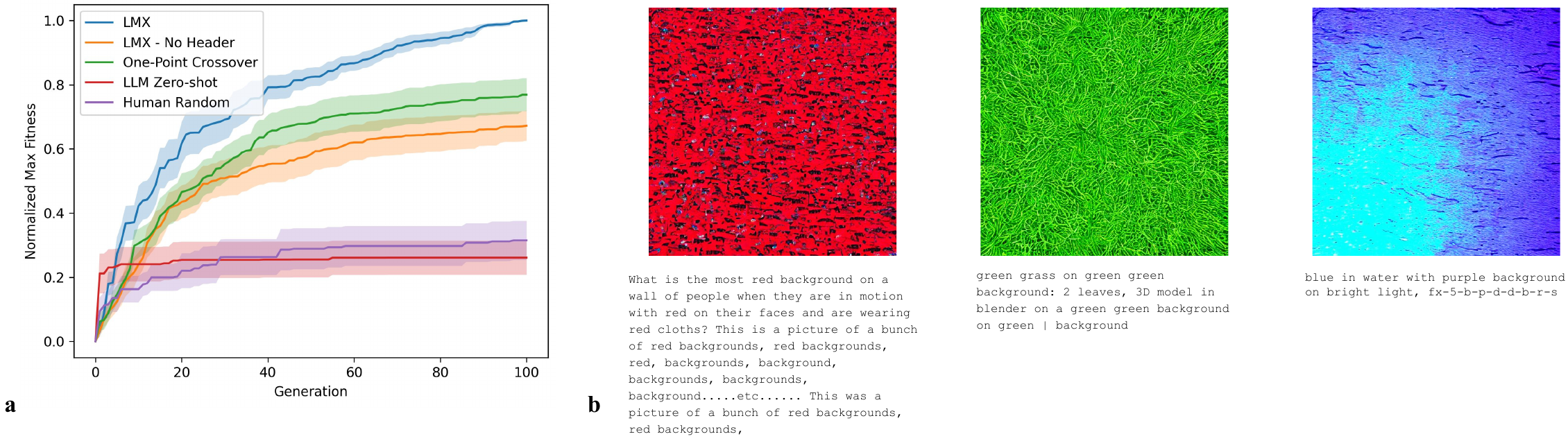}
\caption{\emph{Image generation results.}
(a) Performance aggregated (mean and std. err.) over nine runs (three seeds for each color for each method; normalized to [0, 1] based on the min and max fitness for each each seed) shows that LMX substantially outperforms the alternatives.
One-point crossover is a strong baseline, with performance statistically similar to the LMX - No Header ablation.
The zero-shot baseline quickly stagnates, as it is unable to iteratively refine it's initial solutions; even Human Random solutions eventually outperform it, as they have greater diversity.
(b) The highest-fitness prompts and corresponding images of LMX for each color all include the word ``background'', but vary in the length and detailed content, highlighting LMX's ability to discover diverse, non-obvious solutions.
}
\label{fig:sdx_results}
\end{figure}

Figure~\ref{fig:sdx_results}a shows performance aggregated over nine runs (three seeds for each color for each method; normalized to [0, 1] based on the min and max fitness for each each seed; mean and std. err. shown) shows that LMX substantially outperforms the alternatives.
One-point crossover is a strong baseline, with performance statistically similar to the `no header' ablation, supporting the idea that the ability to naturally incorporate natural language problem specifications is a key advantage of LMX over classical EAs.
The zero-shot baseline quickly stagnates, as it is unable to iteratively refine it's initial solutions; even randomly-selected human prompts eventually outperform it, as they have greater diversity; both these baselines far underperform the evolutionary methods.
So, overall, it is the combination of evolution with the native linguistic capacity of LLMs that makes LMX excel.
Figure~\ref{fig:sdx_results}b shows the highest-fitness prompts and corresponding images of LMX for each color.
All three images have clearly optimized for the target color.
All three prompts include the word ``background'', but vary in the length and kind of detailed content, highlighting LMX's ability to discover diverse, non-obvious solutions.
The conclusion is that LMX can enable sensible
evolution of images.

\subsection{LMX with Python Sodaracers} \label{sec:sodaracer}

Finally, to explore whether LMX can generate variation in code, we apply LMX to evolving Python programs in the Sodarace environment from \citet{lehman2023evolution}, which also explored evolving Python programs with LLMs (we leverage the OpenELM implementation of sodarace \cite{bradley2023openelm}). Sodarace is a 2D simulation of robots with arbitrary morphology constructed from Python functions (the genotype) which output a dictionary specifying joints and muscles, and how they are connected. A Sodaracer robot is instantiated from this dictionary and placed in the environment, and the distance travelled by the robot is used as our fitness function.

We evolve these programs with MAP-Elites \cite{mouret:arxiv15}, using the distance travelled by the generated Sodaracers in a simulation as the fitness and the morpology of the Sodaracer (height, width, and mass) as the dimensions of the behavior space (as in \citet{lehman2023evolution}). We explore the effect on evolution from varying the number of parents that LMX uses to generate offspring (from one to three parents).

Seven pre-existing Sodarace programs were chosen as seeds (details in appendix \ref{appendix:python}). To initialize the population for evolution, LMX was prompted across combinations of these seeds as parents. We randomize the order of seeds for each application of LMX, to control for variance in results from the order of programs in the prompt. The programs were all given the same Python function signature \texttt{make\_walker():} and then concatenated together in the prompt. Note that we begin each completion with this function signature to improve performance (experiments where the LLM prompt did not end with the function signature performed worse; see appendix \ref{appendix:python}). The LLM output is then interpreted as a potential offspring Python program, to be evaluated in the Sodarace environment. 

During evolution steps, we use the same procedure, but randomly select populated niches in the map to select from to build the prompt (as many niches are sampled as parents for each separate treatment), and choose the fittest individual in each niche. We experiment with three different-sized LLMs from the Salesforce CodeGen suite \cite{nijkamp2022codegen}, a set of models trained on a large dataset of code in many languages, including Python.

We perform 10,000 evolutionary iterations (corresponding to 10,000 outputs from the language model, not all of which are valid programs) using 500 initialization iterations. We evaluate the performance of each experimental treatment by computing the percentage of valid programs, number of niches filled at the end of evolution, and the QD score at the end of evolution.


\begin{figure*}
    \centering
    \includegraphics[width=1.0\textwidth]{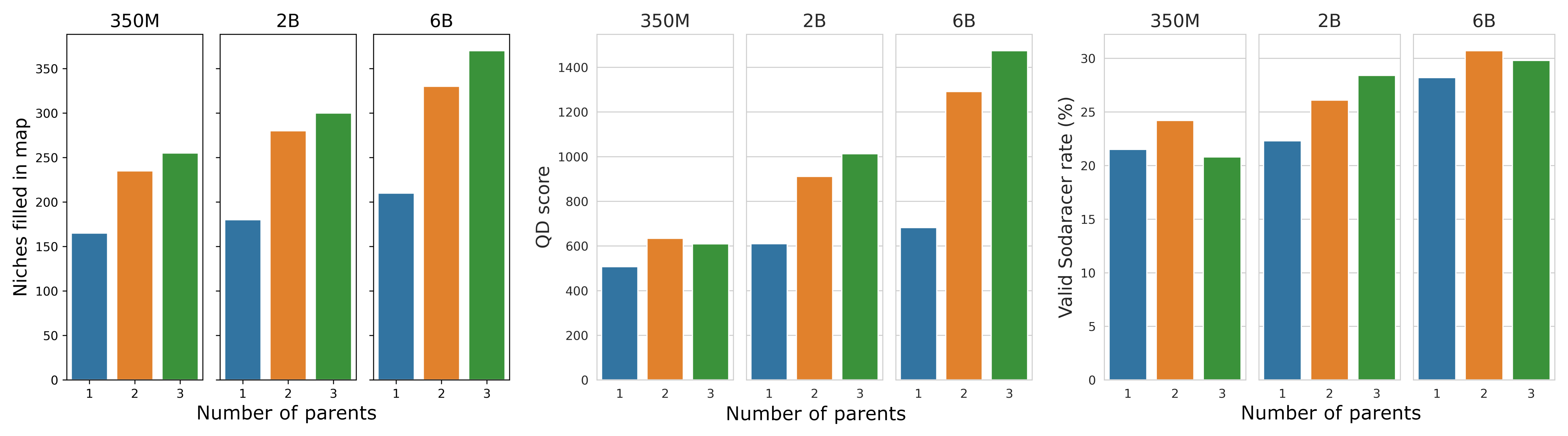}
    \caption{\emph{Sodaracer results}. We show the results for varying numbers of parents in the LLM prompt and across LLM scale. (left) Number of niches filled in MAP-Elites. (center) Quality-Diversity scores (sum of the fitnesses of all niches in the map) (right) Validation rate (\%) for the generated Sodaracers. LMX generally benefits from more examples in its prompt, is able to produce reasonable variation, and often creates valid Sodarace mutations, highlighting its promise for evolving code.}
    \label{fig:sodaracers}
\end{figure*}

The results from these experiments are shown in Figure~\ref{fig:sodaracers}, showing that as the number of parents in the prompt increases, the diversity of offspring generally increases, as measured by the number of niches filled and the QD score (This effect is even more dramatic in the experiments where the LLM prompt did not end with the function signature---A single parent yields no valid offspring (see Appendix Figure~\ref{fig:sodaracers_unforced})). 

Furthermore, a significant proportion of generated offspring are valid Sodaracers (roughly 30\% with the 6B model), highlighting the potential for evolution. Experiments with a single seed in the prompt can be viewed as a simple mutation operator (a different approach to the same end in \citet{lehman2023evolution}). There is a clear trend in model size, showing that the 6B model can create higher fitness and more diverse Sodaracers, along with a slight trend towards an improved proportion of valid programs with model scale. These results therefore demonstrate the promise of LMX for evolving non-trivial Python code.

\section{What Makes LMX Effective?}
\label{sec:what_makes_lmx_effective}

The breadth of experiments in Section~\ref{sec:experiments} show how LMX can serve as a simple and general method for evolution across a range of domains.
This section presents some perspectives on where the effectiveness of LMX could come from, including its connection to EDAs and how it could serve as a starting point for more powerful future algorithms.

\subsection{Connection to EDAs}
\label{sec:eda}

An EDA constructs an \emph{explicit} probabilistic distribution $D$ fit to the parent set $\{x_1, \ldots, x_M\}$, and samples child solutions $x$ from $D$ \cite{hauschild2011introduction, larranaga2001estimation}.
In contrast, a standard GA generates children by sampling from an \emph{implicit} conditional probability distribution $p_g(x \mid [x_1, \ldots, x_M])$ induced by the process of randomly sampling parents and applying a stochastic reproduction operator $g$ (e.g., a crossover operator; Eq.~\ref{eq:variation_operator}).
LMX occupies an intermediate level of explicitness: The conditional distribution induced by feeding the parent prompt into the LLM is \emph{explicit} in that it yields a series of probability distributions over tokens, and the probability of any output sequence can be directly computed, but is \emph{implicit} in the sense that the internal workings of the distribution are encoded opaquely within the millions or billions of parameters and activations of the LLM for a given prompt.

Whatever the level of explicitness, LMX acts like an EDA in that it builds a probabilistic model of parents, from which children are then sampled.
This connection is most clear when LMX takes as input the full population of $n$ potential parents.
Let $S_n$ be a selection operator that refines a collection of $N > n$ candidates down to $n$ (as in Line 13 of Alg.~\ref{alg:lmx}).
\begin{theorem}[eda representation]
    LMX and $S_n$ are sufficient operators to define an EDA.
\end{theorem}
\begin{proof}
    Let $P_t$ denote the current population at iteration $t$ (as when entering the loop at Line 5 in Alg.~\ref{alg:lmx}).
    Then,
    \begin{equation}
        P_{t+1} = S_n \circ \{x_i \sim \textrm{LMX}(P_t)\}_{i=1}^{N>n}
    \end{equation}
    denotes an algorithm (akin to the loop in Alg.~\ref{alg:lmx}) in which at each iteration the next population is constructed by sampling $N$ candidates from LMX conditioned on all of $P_t$ and then refining down to $n$ candidates via $S_n$.
    Using Eq.~\ref{eq:lmx},
    \begin{align}
            P_{t+1} &= S_n \circ \{x_i \sim \psi(\textrm{LLM}(\phi(P_t)))\}_{i=1}^{N>n}\\
            &= S_n \circ \psi \circ \beta_{N} \circ \textrm{LLM}_o \circ \phi(P_t),
    \end{align}
    where $\beta_N$ is the autoregressive sampling operation (Eq.~\ref{eq:llm_autoregression}) applied multiple times to generate $N$ candidates.
    Rotating the recursive composition to the left yields
    \begin{equation}
        D_{t+1} = \textrm{LLM}_o \circ \phi \circ S_n \circ \psi \circ \beta_N(D_t),
    \end{equation}
    where $D_t$ defines the distribution (i.e., model) that candidates are sampled from at iteration $t$.
    Then, $\alpha = \textrm{LLM}_o \circ \phi$ is a \emph{model-building operator} called only once per iteration that constructs a probabilistic model from a set of solutions, and $\beta = \psi \circ \beta_N$ is a \emph{sampling operator} that samples new solutions from a model.
    So, along with the \emph{selection operator} $S = S_n$, we have
    \begin{equation}
        D_{t+1} = \alpha \circ S \circ \beta(D_t),
    \end{equation}
    which is the functional form of an EDA.
    \end{proof}

The key design feature of an EDA is the class of distributions $\mathcal{D}$ to which $D$ belongs.
This class $\mathcal{D}$ can range from simple univariate distributions \cite{baluja:94, harik1999compact} to more complex models like Bayesian networks \cite{pelikan1999boa, pelikan2005hierarchical}.
What is the class $\mathcal{D}_\textrm{LM}$ from which LMX constructs parent distributions?
Due to its in-context learning capabilities \cite{xie:iclr22,rubin:arxiv21}, the LLM can be seen as attempting to infer underlying distribution of parents in the prompt, and to generate continuations accordingly.
By concatenating parents in a random order, the implicit signal to the LLM is that the list is unordered.
The LLM may notice some accidental patterns in the order, but, as the number of parents increases, e.g., when LMX processes the full population as in the EDA above, the significance of such spurious patterns diminishes and a well-trained LLM is more likely to perceive the order as random.
These parents are text-based objects that must have been sampled from some ground truth distribution $D^*$, and thus the LLM's highest-probability action is to keep sampling objects from $D^*$ as it generates output.
In other words $\mathcal{D}_\textrm{LM}$ consists of distributions of \emph{objects that are found in sets that might appear in the universe} of data from which the dataset used to train the LLM was drawn.
An ideal EDA would select the most probable $D = D^*_{\textrm{EDA}} \in \mathcal{D}_\textrm{LM}$ based on the parent set $\{x_1, \dots, x_k\}$. E.g.,
\begin{equation}
    D^*_{\textrm{EDA}} = \argmax_{D \in \mathcal{D}_\textrm{LM}} p(D) \prod_{i=1}^k p(x_i \mid D),
\end{equation}
where $p(D)$ is the prior probability of $D$ in $\mathcal{D}_\textrm{LM}$.
As the LLM becomes a better and better in-context learner, it becomes better able to detect subtler patterns within a prompt of randomly-ordered concatenated parents, and thus
\begin{equation}
    \textrm{LMX}(x_1, \ldots, x_k) \approx D^*_\textrm{EDA}.
\end{equation}
Note that the left side depends on an ordered list of parents, while the right side has removed this dependency on order.

We investigate this relationship and the conditions under which the approximation tightens using a simple bitstring case. Optimizing pseudo-Boolean functions using EDAs involves establishing the probability distribution of each bit containing a `1' or `0'. The Univariate Marginal Distribution Algorithm \cite{muhlenbein1996recombination}, the prototypical EDA, samples $\lambda$ individuals each iteration, choosing the best $\mu$. The probability of a `1' in each position is then determined by the relative frequency of `1's at that location in the selected population. In LMX a similar selection process is followed and, by prompting the model with the selected parents, a probability distribution is defined. 

Despite the implicit definition in LMX, the probability distributions produced by LMX and an EDA can be directly compared. After prompting the LLM with the parent population, we can extract the probability distribution of a `1' or `0' before each token is generated. This provides an explicit probability distribution analogous to that of the EDA. In this way we can test the hypothesis that LMX approximates an EDA more closely as the size of the parent population increases. We examine the similarity of distributions with increasing populations in a six-bit case with the following procedure:
\begin{enumerate}
\item For each bit in the string, the probability of it being a $1$ or $0$ is drawn uniformly at random from $[0, 1]$.
\item A set of parents is generated according to the distribution established in the previous step.  
\item Given this set of parents the mean absolute difference in the probability of a 0 for each gene between the resulting EDA and LMX distributions is calculated.
\item The entire experiment is repeated with a different initial probability distribution.
\end{enumerate}

\begin{figure}[]
    \centering
    \resizebox{\textwidth}{!}{\input{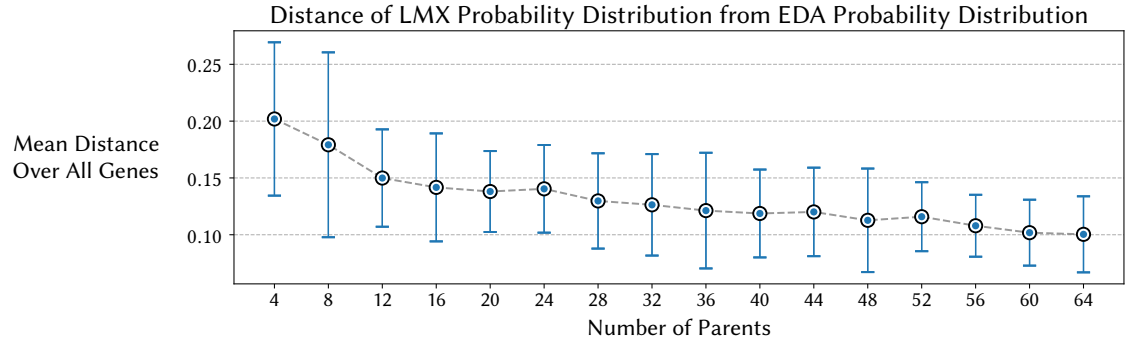}}
    \caption{\emph{LMX and EDA Probability Distributions Across Different Sized Parent Sets.} The average difference in gene probabilities predicted by LMX and EDA approaches 0 across various parent set sizes in a population of bit strings. Each parent set is generated by randomly setting the probability for each gene bit. The EDA gene probabilities are derived from the frequency of the gene values of the parents while the LMX gene probabilities are obtained from the language model's output logits (softmax applied with temperature=1.0). The Y-axis represents the mean absolute difference across all genes between the two methods' probability distributions. Error bars indicate the standard deviation over 20 experiments. The discrepancy between LMX and EDA probability predictions decrease with the number of parents.}
    \label{fig:eda_compare}
\end{figure}

When we examine the difference between the EDA and LMX distributions with an increasing number of parents (Figure \ref{fig:eda_compare}), we find that indeed the disparity between the two distributions diminishes as the number of parents increases, i.e., LMX becomes more similar to the EDA.

Though a faithful application of an EDA may include the full parent population in each parent prompt, the experiments in Section~\ref{sec:experiments} save compute by sampling only a small number of parents. Nonetheless, by comparing LMX to EDAs it may be possible to analyze the optimization behavior of LMX  \cite{krejca2020theory} (e.g.,\ global convergence analysis \cite{zhang2004convergence}), as discussed in Section~\ref{sec:discussion}. 
Note that, as we are using a causal LLM, probabilities of each bit are not technically independent, but rather conditional on the previously generated bits in the genome as well as on the order of the parents.
This nuanced scenario is also characteristic of more sophisticated EDAs that incorporate conditional dependencies \cite{shakya2012review}.
Despite this confounding factor, it is clear that with a larger number of parents both approaches converge toward the same distribution -- and this connection to EDAs may help to explain why LMX is effective as an off-the-shelf genetic operator across a wide range of domains.

\subsection{Universality of LMX}
\label{appendix:universality}
Section \ref{sec:eda} highlighted the connection between LMX and EDAs. This section explores another property of LMX, its theoretical universality (i.e.\ its ability in theory to express any genetic operator). With a sufficiently expressive class of model, such as Bayesian networks \cite{pelikan1999boa, pelikan2005hierarchical}, EDAs 
can approximate any candidate distribution as the size of the parent set increases \cite{zhang2004convergence}.
Not only can LMX sample from distributions represented by an EDA, but it can in principle sample from any conditional probability distribution, making it universal in the space of genetic operators, even with small parent sets.
Recent theoretical work has shown how crossover of large neural networks can yield universal approximation of reproduction distributions \cite{meyerson2022simple}.
LMX also achieves theoretical universal approximation via large neural networks, but by feeding parents directly into the LLM, instead of crossing-over weights.
If modification (e.g., fine-tuning) of LLM weights is permitted, this result follows naturally from the universal approximation ability of NNs \cite{cybenko1989approximation, hornik1989multilayer, kolmogorov1957representation} (note that this property also applies in the single-parent case for mutation-based evolution through LLMs \cite{lehman2023evolution}):
\begin{theorem}[weight-based universality]
    \label{thm:weight_universality}
    For any genetic operator $g$ on candidate space $\mathcal{X}$ and $\epsilon > 0$, if $\phi: 2^\mathcal{X} \to V^*$ is injective, and $\psi: V^* \to \mathcal{X}$ is surjective, then $\exists$ an LLM s.t. for all parent sets $X$ of $g$ and children $x$
    \begin{equation}
        \Big\lvert \ \Pr\big[x \ \big| \ g(X)\big] - \Pr\big[x \ \big| \ \textrm{LMX}(X)\big] \ \Big\rvert < \epsilon.
    \end{equation}
\end{theorem}
\begin{proof}
    $\textrm{LMX}(X) = \psi(\textrm{LLM}(\phi(X))).$
    Since $\phi$ is injective, for all parent sets $X$, $\phi(X) = s_X$ is unique.
    Since $\psi$ is surjective, $\forall \ x \ \in \mathcal{X}$, $\exists \ s_x$ s.t. $\psi(s_x) = x$.
    Let $S_x = \{s_x : \psi(s_x) = x\}$.
    It suffices to find an LLM with weights such that
    \begin{equation}
        \Big| \ \Pr\big[x \ \big| \ g(X)\big] - \Pr\big[s_x \in S_x \ \big| \ \textrm{LLM}(s_X)\big] \ \Big| < \epsilon.
    \end{equation}
    The existence of such an LLM exists follows from the universal approximation capability of transformers \cite{yun2019transformers}.
\end{proof}
However, when coupled with external memory, \emph{existing fixed pre-trained LLMs} today, e.g., Flan-U-PaLM 540B \cite{chung2022scaling}, have been shown to implement universal Turing machines (UTMs) \cite{giannou2023looped, schuurmans2023memory}, implying that universality can be achieved through effective prompting schemes, without altering LLM weights:
\begin{theorem}[prompt-based universality]
    \label{thm:prompt_universality}
    For any genetic operator $g$ on candidate space $\mathcal{X}$ and $\epsilon > 0$, if the LLM is a UTM and $\psi$ is surjective, then $\exists \ \phi$ s.t. for all parent sets $X$ of $g$ and children $x$
    \begin{equation}
        \Big\lvert \ \Pr\big[x \ \big| \ g(X)\big] - \Pr\big[x \ \big| \ \textrm{LMX}(X)\big] \ \Big\rvert < \epsilon.        
    \end{equation}

\end{theorem}
\begin{proof}
    
    As in Thm.~\ref{thm:weight_universality}, $\textrm{LMX}(X) = \psi(\textrm{LLM}(\phi(X)))$, and since $\psi$ is surjective, it suffices to find $\phi$ such that
    \begin{equation}
        \Big| \ \Pr\big[x \ \big| \ g(X)\big] - \Pr\big[s_x \in S_x \ \big| \ \textrm{LLM}(\phi(X))\big] \ \Big| < \epsilon.        
    \end{equation}

    Since the LLM is a UTM, we can choose $\phi$ to implement a program with argument $X$.
    Such a program exists, since $g$ implements its behavior up to the equivalence classes of $\psi$.
\end{proof}

This property suggests that the power of LMX is not limited to the randomly-ordered-parent-concatenation-based crossover demonstrated in this paper, but could be used to produce (manually or automatically) crossover behavior optimized for specific tasks, e.g., through prompt-engineering.
This ability to achieve arbitrarily complex and diverse reproductive behavior within a single framework gives LMX a potential advantage over genetic operators that are hand-designed for different tasks: In theory, LMX can represent all such operators (especially if they appear in the dataset used to train the LLM).
The generative distributions in the experiments in this paper are limited to ones induced by simple concatenation of parents, but the underlying universality of the LMX method in general provides further explanation for how it can be an effective generic operator across such a wide range of domains, as was demonstrated in Section~\ref{sec:experiments}.

Finally, LMX can be used to construct \emph{expressive encodings}, which places it on a common theoretical footing alongside other popular and powerful evolutionary substrates including genetic programming and neuroevolution \cite{meyerson2022simple}.
The power of expressive encodings comes from the fact that, as in natural evolution, the complexity of reproduction (here the LLM) is found in the genome itself, and largely shared across individuals in a population.
Suppose we have genotypes $x \in \mathcal{X}$ and the original encoding $E$ is surjective, i.e., every phenotype has some corresponding genotype.
Now, pack the complexity of the system into the genotype by instead considering genotypes of the form $(x, LLM, \phi, \psi)$:
\begin{theorem}[lmx-based expressive encoding]
    $E_\textrm{LMX}((x, \textrm{LLM}, \phi, \psi)) = E(x)$ is an expressive encoding.
\end{theorem}
\begin{proof}
Consider the following recombination operator:
\begin{align}
    g\Big((x_1, \textrm{LLM}_1, \phi_1, \psi_1), (x_2, \textrm{LLM}_2, \phi_2, \psi_2)\Big) &= \Big(x \sim \psi_1(\textrm{LLM}_1(\phi_1(x_1, x_2))), \textrm{LLM}_1, \phi_1, \psi_1\Big)\\
    &= \Big(x \sim \textrm{LMX}(x_1, x_2), \textrm{LLM}_1, \phi_1, \psi_1\Big).
\end{align}
This operator takes a constant number of parents and has constant description length modulo its arguments, so $g$ is a \emph{simple genetic operator} (SGO) \cite{meyerson2022simple}, and, since $E$ is surjective, by either Thm~\ref{thm:weight_universality} or Thm~\ref{thm:prompt_universality} for any density $\mu$ over phenotypes, we can find 
$x_1$, $x_2, \textrm{LLM}_1$, $\phi_1$ and $\psi_1$ so that $\mu$ is approximated arbitrarily closely by
\begin{equation}
    E_\textrm{LMX}(g((x_1, \textrm{LLM}_1, \phi_1, \psi_1), (x_2, \textrm{LLM}_2, \phi_2, \psi_2))).
\end{equation}
Thus, $E_\textrm{LMX}$ satisfies the definition of an expressive encoding.
\end{proof}
Although for simplicity, in the SGO in this proof the LLM, $\phi$, and $\psi$ are not altered, in general they could be, and this view suggests that simultaneously evolving the solution $x$ along with the LLM and the prompting mechanism (i.e., $\phi$ and $\psi$) could be a powerful paradigm for more open-ended systems.


\subsection{Biasing the LMX Operator through Parental Ordering}
\label{sec:parent_ordering}

As an example of how LMX could move beyond its instantiation as an LLM-based EDA, this section investigates the effect of the ordering of parents within the parent prompt. 
EDAs typically assume an unordered distribution, yet the inherent input ordering in autoregressive models creates a unique opportunity for directing the operator's output. Acting as pattern completion engines, these models provide a pathway to guide the sampling process through directional cues. An ascending input prompt produces ascending output. Consequently, an input arranged in ascending fitness order prompts the model to generate output that follows an ascending fitness trend.

\begin{figure}
    \centering
    \resizebox{\textwidth}{!}{\input{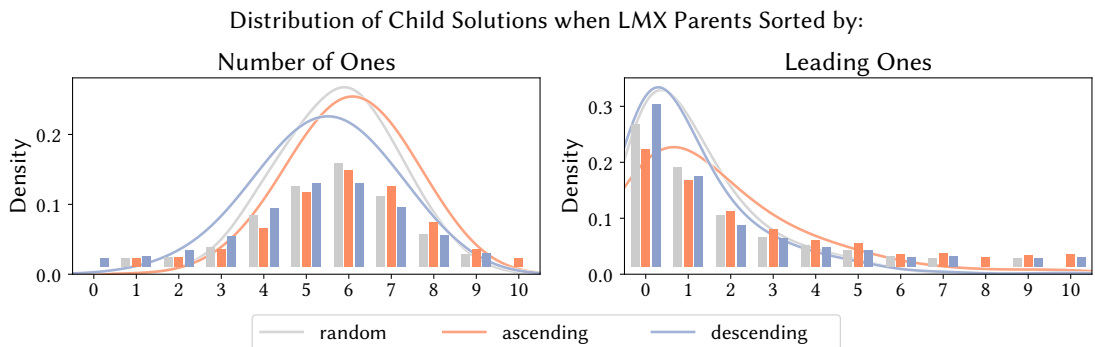}} 
    \caption{\emph{Impact of parental sorting on the distribution of offspring produced using the LMX operator.} The same set of parents, represented as bitstrings, was sorted in ascending, descending, or random order based on one of two fitness criteria: number of `ones' (Left) and leading `ones' (Right). Kernel Density Estimation (KDE) curves reflect the overall offspring distribution trends, bars represent number of individual offspring. The offspring distribution patterns mirror the parental sorting order, underscoring the influence of parent order on the LMX operator's output. Results are cumulative over 100 experiments with 10 children produced in each experiment.}
    \label{fig:order}
\end{figure}

Figure \ref{fig:order} illustrates this biasing technique using the LMX operator in the context of the one-max problem. When randomly generated parents are arranged in ascending order of the 'ones' count, the resulting offspring distribution exhibits a clear skew towards higher fitness (left) and vice versa. The Kernel Density Estimation (KDE) curves clearly represent how offspring distribution is influenced by the sorting order of the parents. The precise counts indicate that for scores of 5 or less from a 10-bit string, the offspring are more likely to originate from parents sorted in descending order than ascending. Conversely, scores of 6 or more tend to come from parents sorted in ascending order vs. descending. Ordering based on the number of leading ones yields a different bias consistent with the ordering (right). 

Subsequent work following the arxiv release of the LMX paper~\cite{fernando2023promptbreeder, yang2023large} performed extensive experimentation with different orderings of parents in more complex natural language domains, and came to similar conclusions about the importance of ordering. Later work, which used an LMX-like approach to discover new solutions to mathematical problems~\cite{funsearch} also sorted parents by fitness before recombination.
These results are suggestive of the versatility of this biasing technique; a variety of sorting strategies could be employed to cater to specific objectives. For example, including auxiliary helper objectives and rankings such as those used in multiobjectivization \cite{jensen2004helper, mouret2011novelty, schmidt2011age}, genotypic niching \cite{niching,fitshare}, novelty \cite{ns-ecj,lehman2011evolving}, or quality-diversity \cite{mouret:arxiv15, cully2017quality} could bias LMX to generate solutions that differ from those previous discovered or that exhibit specific attributes (see Section~\ref{sec:discussion} for a larger discussion).
Though an important research direction worth exploration, this paper has presented LMX at it's most basic and fundamental -- in the experiments in Section~\ref{sec:experiments} all orderings are random.


\section{Limitations}
\label{sec:limitations}

The above sections have illustrated the advantages of LMX.
Here, we discuss limitations that may arise in practice for LMX in its current form.

The experiments in this paper each made use of GPU compute to perform LMX.
Although there is a fast-progressing effort to enable efficient LLM inference on CPU \cite{shen2023efficient, zhang2024nomad}, there is still a benefit to GPU access, which may make LMX less accessible to users without such resources.
There is also the issue that the LLM can generate ``invalid'' outputs that cannot be parsed as solutions, while in traditional EAs validity can be formally guaranteed by carefully-designed genetic operators.
If there is a high proportion of invalid outputs for a certain domain, e.g., due to its complexity or distance from the LLMs training data, then the computational cost of creating the next generation will increase.

Relatedly, if an application domain is especially far from the LLM's training set, e.g., far from any kind of text found on the internet, then choosing a method of population initialization becomes particularly important.
Asking the LLM to sample text unconditionally is certainly not an option, since the output space of the LLM is of the astronomical size $V^{T_\text{out}}$, where $V$ is the LLM's vocab size ($\approx$ 30,000 for the LLMs in this paper), and $T_\text{out}$ is its maximum number of output tokens ($>100$ in this paper).
In the experiments in this paper, the initial population consisted of uniform random samples from the phenotype space (Section~\ref{sec:binary}), random samples from an existing dataset of human-designed examples (Sections~\ref{sec:symbolicregression} and \ref{sec:image_generation}), or a small fixed set of seed examples (Sections~\ref{sec:sentiment} and \ref{sec:sodaracer}).
If a domain is so complex and novel that the above methods are challenging to implement, a fallback approach could be to use a traditional EA to create an initial population and then convert that population to text.

LMX in its current form is also limited by the size of the LLM context window, which, for the LLMs used in this paper, is on the order of thousands, meaning it will not be able to generate solutions larger than this.
EAs have been shown to be effective in search spaces with millions and even billions of variables \cite{deb2016breaking, chicano2017optimizing}.
LMX will not be effective in such spaces unless it is possible to break the problem down into appropriately-sized subproblems, possibly relying on a traditonal EA to discover this modularization.
However, even for relatively small search spaces, if LMX is not behaving as expected on a particular problem, it may be difficult to find the precise cause of the issue, due to the opacity of LLMs, whereas in traditional EAs it could be straightforward to reason about the kinds of variation a genetic operator is likely to produce.
This opacity could also obscure undesirable biases hidden in the model, which could be especially concerning if the application has societal implications.
If they are a concern, the biases of a particular LLM can be investigated independently of the LLMs role in LMX.
Promising progress is being made on interpretability of LLMs \cite{conmy2023towards, singh2024rethinking}, and though it is still in the early stages, if successful, such methods could be readily applied to diagnose issues in LMX and used to identify areas for improvement.

Finally, LMX alone does not allow us to escape common population dynamics concerns that plague traditional EAs, such as premature convergence, diversity maintenance, and other exploration/exploitation pitfalls.
LMX does not solve these grand challenges of EAs, rather it provides a generic genetic operator that allows practitioners to quickly set up algorithms to search diverse spaces in a manner aligned with the generative nature of LLMs, i.e. aligned with the way humans produce text. 

\section{Discussion and Conclusions}
\label{sec:discussion}

As a flexible and easy-to-use genetic operator, LMX provides a way for EA practitioners to take advantage of the recent revolution in large language models.
The experiments in this paper tackle a broad range of potential applications, across
equations, plain-text sentences, images, and code, leveraging the burgeoning ecosystem of open-source neural networks as means of generating variation, crossing modalities, and measuring both fitness and diversity. 
LMX could even be used for continuous optimization, as suggested by how it tunes floating-point constants (Figure~\ref{fig:sr_dynamics}).

There is much room for future work. The experiments focused on breadth rather than depth, and it is possible
that with further effort LMX could enable state-of-the-art results, e.g., in symbolic regression. One important direction is to explore the dependence of LMX's performance on qualities of the underlying LLM; the ability of LMX to suggest relevant variation in a particular domain is likely dependent on the LLM's training data (along with its size and how well it was trained). For example, the expectation is that if the type of text chosen for evolution (e.g.\ code in a very new programming language) is not well-represented in the training distribution of a particular LLM, then LMX that relies on that LLM will likely perform poorly.
Larger LLMs generally have been trained on larger and broader data sets, so are more likely to be comfortable processing data from a given domain.
If larger models are infeasible to run, for any given model size there may be a trade-off between more generic models, which are likely to perform decently in most domains, and more specialized ones.
The discrepancy between Pythia and Galactica in Section~\ref{sec:symbolicregression} is evidence for this trade-off, though more would need to be done to see how widespread it is in other domains.
More capable models would be helpful both on the LLM side, e.g., for maintaining coherence of generated solutions, and on the evaluation side, e.g., improved sentiment and visual aesthetic evaluation models should lead to improved results in Sections~\ref{sec:sentiment} and \ref{sec:image_generation}.
As LLMs get more and more reliable, it may be possible to develop convergence bounds of EAs driven by LMX.
Such theoretical work could be based on establishing assumptions on the training data distribution of LLMs, and then linking methods from machine learning theory with those from EA theory, with implications for validity and variation in a single step of LMX used to bound convergence for the a full run of the algorithm.
For example, the level of stochasticity in LMX can be controlled by the softmax temperature parameter, which can be seen as analogous to a mutation rate parameter in a traditional EA.
The connection to EDAs could enable the applications from EDA theory to this end \cite{zhang2004convergence, krejca2020theory}.
The experiments in this paper relied on base LLMs, but it would be possible to develop variants of LMX that apply to instruction-tuned LLMs \cite{chung2022scaling}, i.e., where the prompt describes explicitly the type of variation behavior desired.
Theoretical progress could be amenable in such a case, by making explicit assumptions that the LLM will follow instructions with some level of reliably.

Another interesting question is whether examples fed into LMX could be chosen more deliberately (e.g.\ only crossing-over similar individuals to get more nuanced variation); preliminary experiments showed some qualitative effect from applying LMX on individuals with similar embeddings (Section~\ref{sec:sentiment}), but require further experimentation to validate.
Such work would benefit from a generalization of the LMX heritability and diversity analysis in Section~\ref{sec:binary} to complex domains where embeddings are the most practical space for computing distances between solutions.
The results in Section~\ref{sec:parent_ordering} indicate that different ways of feeding the parents into the LLM can have a substantial impact on the generative process, giving yet another knob to control the nuance of variation.
One natural future direction is to explore whether there is 
benefit from combining the recombination capability of LMX
with the mutation operators (either prompt-based or diff-model-based) explored in ELM \cite{lehman2023evolution}. Of further interest is the possibility for self-improvement of LMX (as in ELM), through fine-tuning the model on successful examples of variation in a domain. A final intriguing possibility is the use of LMX for interactive evolution, e.g.\ to interactively evolve sentences, code, or images \cite{secretan2008picbreeder,bontrager2018deep}.
More generally, beyond the GA framework used in this paper, LMX could be integrated into many other EA frameworks, such as EDAs, as suggested in Secton~\ref{sec:what_makes_lmx_effective}; it is possible that within the dynamics of other frameworks new synergies would be uncovered.
Sections~\ref{sec:sentiment} and \ref{sec:sodaracer} investigated applying LMX in a QD setting; Section~\ref{sec:symbolicregression} applied LMX in a multi-objective setting, but with only an implicit bias towards parsimony governed by the max output length of the LLM.
We would expect even more interesting dynamics to emerge in more sophisticated applications LMX in QD and multi-objective EAs.

While LLMs are computationally expensive, all of the experiments in this paper (with exception of the Python Sodaracer experiment) were conducted either through Google Colab notebooks or on a single GPU; the code to run experiments is surprisingly compact, as the LMX method consists mainly of a simple LLM prompting strategy, and interacting with language and image models has become simple through APIs and libraries such as those provided by HuggingFace. In conclusion, there are likely many creative ways to beneficially combine various models together that this paper leaves unexplored; evolution in general is a powerful and easy-to-implement way to quickly explore such possibilities, and LMX in particular is a promising and simple way of instantiating them.



\bibliographystyle{ACM-Reference-Format}
\bibliography{references}

\pagebreak
\appendix

\section{Binary String Experimental Details}
\label{appendix:binary}

The base LLM used for these experiments is the Pythia-deduped 800M model. For the scaling experiments, different parameter sizes were used (as noted in the text). All models are hosted on Huggingface. These Pythia models are trained by EleutherAI for ongoing research \cite{eleutherai:pythia23}.

Samples from the LLM were set at a maximum of $150$ tokens. 
For these experiments, rather than using the temperature hyperparameter for controlling LLM sampling, the
top-$k$ and top-$p$ hyperparameters are used. 
Top-k restricts the LLM to output only from the $k$ highest-probability tokens. Top-p further restricts the tokens to be the top tokens that cumulatively take up $p$ 
of the probability mass. With mild tuning (i.e. rough hand-tinkering to get a sense of what sampling hyperparameters make sense for this LLM), for all experiments in this section, top-$p$ was set to $0.8$ and
top-$k$ was set to 30.

The evolutionary algorithm used truncation selection of the top 50\% of the population and elitism. The population size was 10, the number of bits in the bit string 10, and the number of generations 10. The one point crossover operator chose two parents at random from the truncated set of parents, performed one point crossover two produce a single child, then flipped each bit in the genome with a probability of $0.1$. 

Tokenizers vary according to the model, for instance $00$ and $111$ may each be a single token. In the binary string experiments to avoid unintended effects from the tokenizer, rather than evolving strings like $0011$ all genomes were pre and post processed and given to the language model with "\_" characters separating each bit (e.g. $0011$ becomes $\_0\_0\_1\_1$) ensuring each bit is a single token.


\subsection{Binary Strings Model Scaling}
\label{appendix:binary_scaling}

In this experiment, the number of parents is fixed to 3, and a range of models from the Pythia suite are applied in the same way as in the variation experiment of section \ref{sec:binary}, i.e.\ to generate variation from randomly-sampled binary strings (although in this experiment they are of length 9 as opposed to length 6). When averaged over 15 randomly-generated parent sets, both the percent of valid offspring and number of novel offspring generally increase with model size (Figure \ref{fig:binary_1}b).

\section{Symbolic Regression Experimental Details}
\label{appendix:sr}

GALACTICA 1.3B was used as the LLM \cite{taylor2022galactica}.
The sampling temperature was set to 0.8, which in initial tests was found to noticeably improve consistency of generated output compared to the default of 1.0.
All other sampling parameters were defaults.

The initial population had 1000 candidates, and population size was set to 50 thereafter.
A larger population could improve performance; here it was limited to 50 for computational efficiency. 
Any generated offspring that was already in the population was immediately discarded without being evaluated.
To prevent stagnation with this relatively small population size, throughout evolution, there was always a 0.05 probability of generating a new candidate directly from the prior set of benchmark expressions (randomly selecting an expression and randomly mapping variables) instead of through LMX.
The benchmark expressions are popular benchmarks, whose python representations were copied from the `deep-symbolic-optimization' GitHub repository (\url{github.com/brendenpetersen/deep-symbolic-optimization/}).

Text length for the LLM was capped at 500 tokens.
Running 5000 generations took around 100hrs.
The vast majority of wall-clock time is spent in the forward pass of the LLM.
This could be reduced considerably through batching offspring generation, which is naturally parallelized.

\section{Modifying Sentiment Experimental Details}
\label{appendix:sentiment}

The LLM used in this experiment for LMX is the 1.4 billion parameter Pythia model, hosted on HuggingFace. As in the binary string experiment, for sampling, top-$p$ was set to $0.8$ and
top-$k$ was set to 30. The max number of tokens generated was set to 128.

\begin{figure}
    \centering
    \includegraphics[width=0.5\columnwidth]{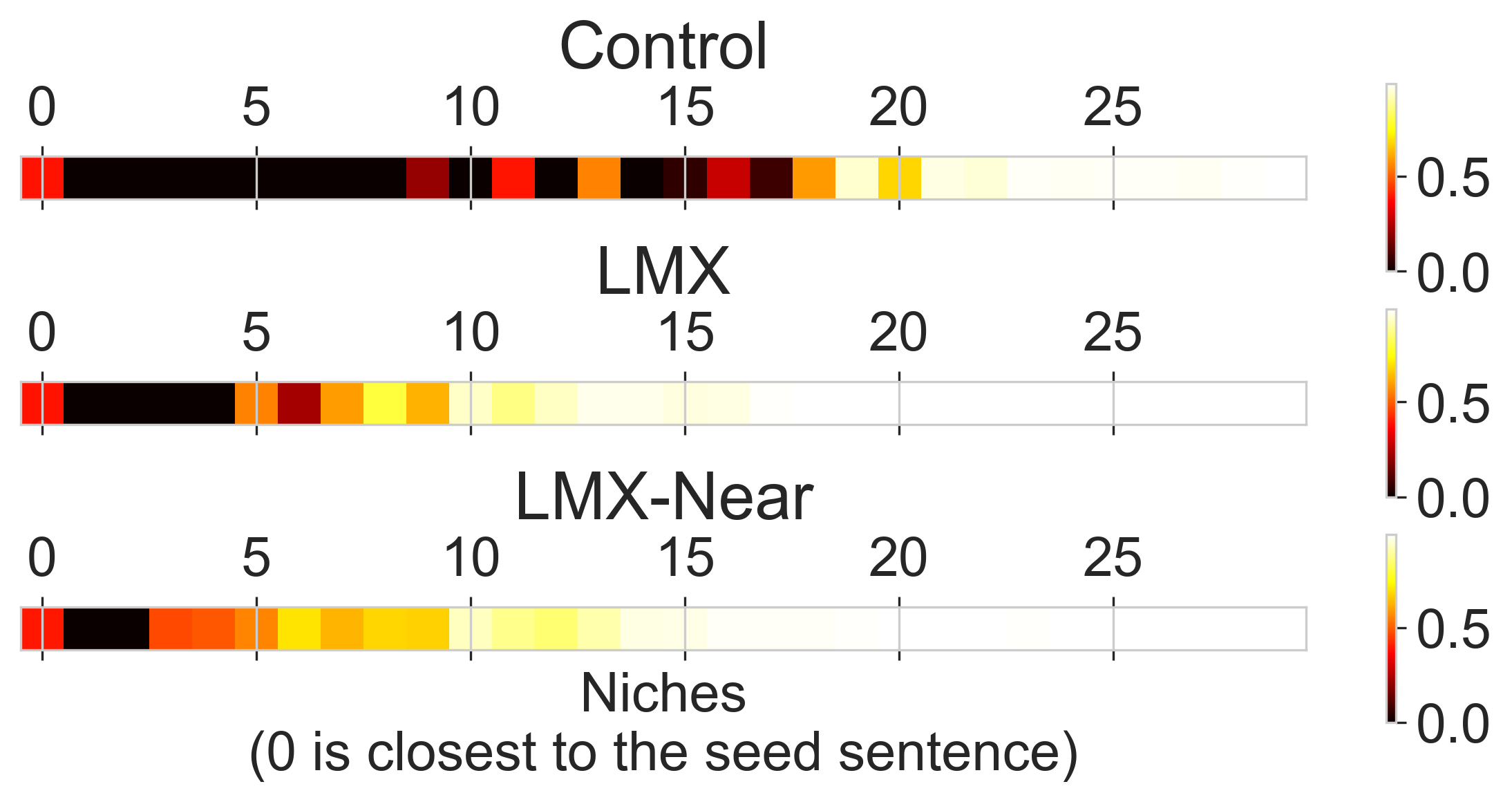}
    \caption{Representative MAP-Elites Heatmaps for Homer Simpson Quote Sentiment. Shown are heatmaps of the final QD maps discovered by MAP-Elites for LMX-Near, LMX, and the baseline control. The left-most grid squares are niches for sentences most like the original quote, while the furthest grid squares are for sentences very far from the original quote (according to the embedding model). Black indicates the map square was not filled, while white indicates maximum fitness (1.0). The control struggles to fill many niches, especially those nearest to the start sentence. LMX-Near performs better than LMX in this case, filling a few extra niches nearer to the start sentence.}
    \label{fig:sentiment_heatmap}
\end{figure}

\begin{figure}
    \centering
    \includegraphics[width=0.7\columnwidth]{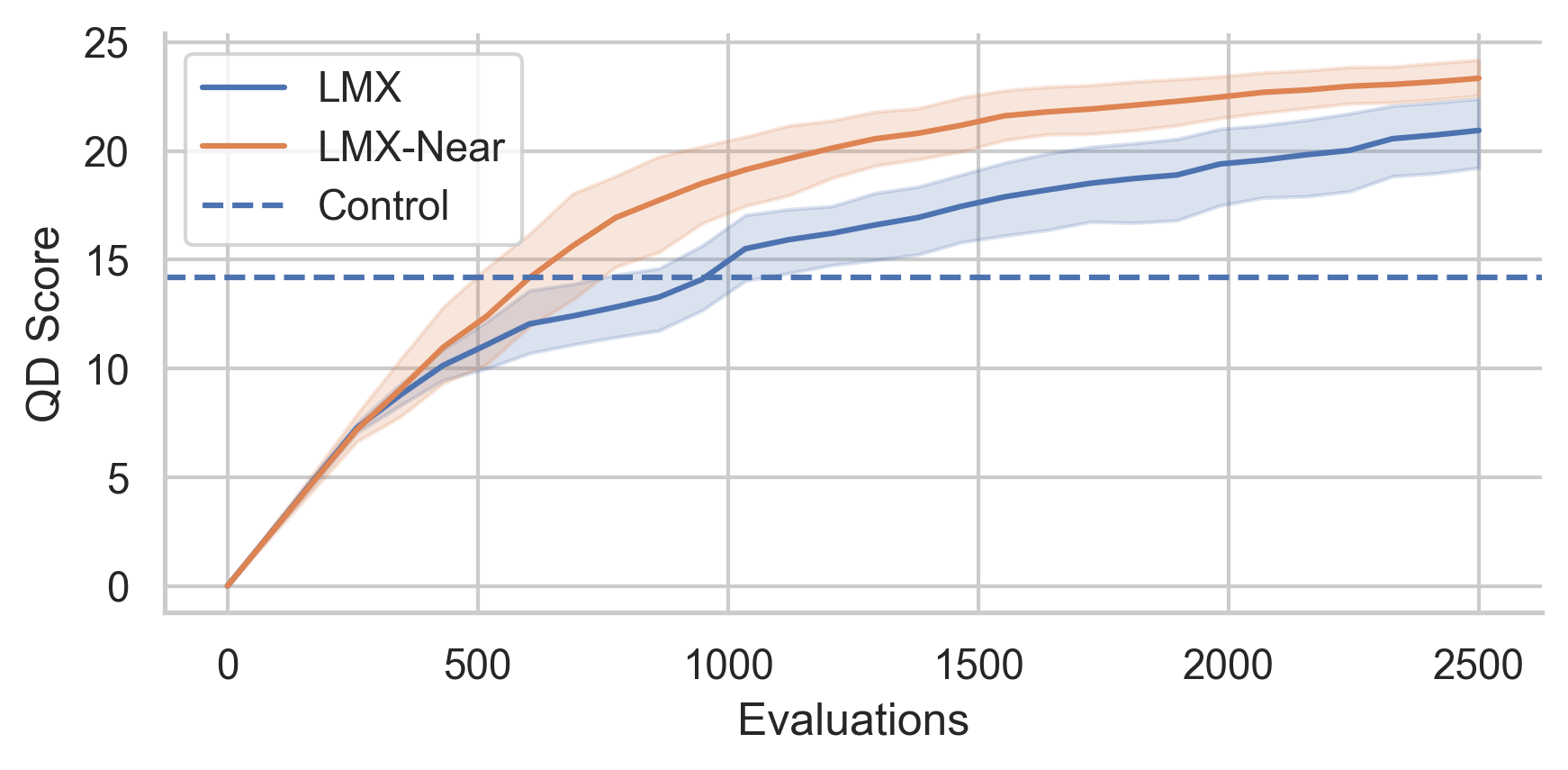}
    \caption{Modifying Gore Vidal Quote Sentiment. The plot compares LMX-Near, LMX, and the baseline control in increasing the positive sentiment of the quote: ``Whenever a friend succeeds, a little something in me dies.'' LMX-Near outperforms LMX significantly, and both significantly outperform the control. Example sentences of such runs are shown in appendix section \ref{appendix:sentiment_results}.}
    \label{fig:sentiment0}
\end{figure}

\begin{figure}
    \centering
    \includegraphics[width=0.7\columnwidth]{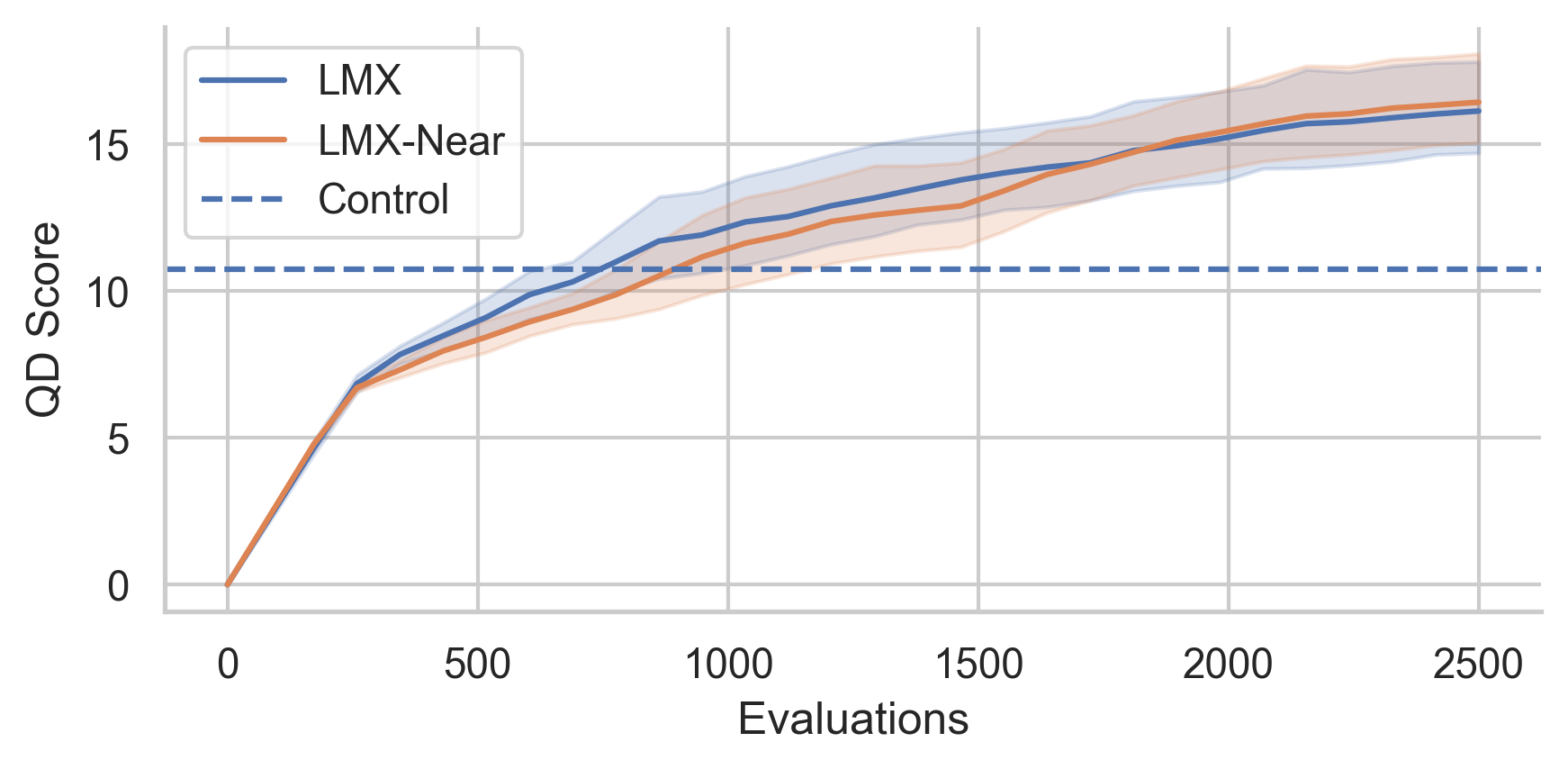}
    \caption{Modifying Woody Allen Quote Sentiment. The plot compares LMX-Near, LMX, and the baseline control in increasing the positive sentiment of the quote: ``Life is divided into the horrible and the miserable.'' LMX and LMX-Near do not perform significantly differently, but both significantly outperform the control. Example sentences of such runs are shown in appendix section \ref{appendix:sentiment_results}.}
    \label{fig:sentiment2}
\end{figure}

Figure \ref{fig:sentiment_heatmap} shows representative heatmaps of final MAP-Elites maps for each treatment for the Homer Simpson quote. Figure \ref{fig:sentiment0} shows fitness plots for the Gore Vidal quote, Figure \ref{fig:sentiment1} shows fitness plots for the Homer Simpson quote, and Figure \ref{fig:sentiment2} shows fitness plots for the Woody Allen quote. Further examples of evolved behavior are shown in appendix section \ref{appendix:sentiment_results}.

\subsection{Additional Positive Sentiment Results}
\label{appendix:sentiment_results}

The full Pareto front for one representative run of modifying the Simpsons quote sentiment (from LMX-Near) is shown in Table~\ref{table:fullpareto}.

\begin{center}
\begin{table}
\begin{tabular}{|c|c|p{5cm}|}
\hline
\textbf{Distance} & \textbf{Positivity} & \textbf{Sentence} \\ \hline \hline
0.00 & 0.021 & Kids, you tried your best and you failed miserably. The lesson is, never try. \\ \hline
0.09 & 0.023 & Kids, you tried your best, but you failed miserably. The lesson is, never try. \\ \hline
0.12 & 0.049 & Kids, you tried your best, but you failed miserably. The lesson is, always try. \\ \hline
0.20 & 0.103 & Kids, you tried your best, but you failed. the lesson is, never stop trying. \\ \hline
0.21 & 0.157 & Kids, you always tried your best and you failed. The lesson is, never stop trying. \\ \hline
0.25 & 0.158 & Kids, you tried, tried your best, but you failed. The lesson is, never stop trying. \\ \hline
0.28 & 0.886 & Kids, you tried your best. The lesson is, you always succeed. \\ \hline
0.36 & 0.901 & Kids, you tried your best. The lesson is, success is guaranteed. \\ \hline
0.42 & 0.938 & Kids, you did your best. The lesson is, you never stop trying. \\ \hline
0.46 & 0.956 & Kids, you went above and beyond. The lesson is, never fail, but always try. \\ \hline
0.50 & 0.961 & Kids, you always succeed, The lesson is never fail, but always try, and as long as you keep trying, you will succeed. \\ \hline
0.56 & 0.964 & Kids, you have proven yourself a winner. The lesson, is, never give up, but always try, and as long as you keep trying, you will succeed. \\ \hline
0.61 & 0.983 & Kids, you're the best ever. The lesson is, the best always wins. \\ \hline
0.72 & 0.987 & Kids, you're the best. You're the best, the best. The best. \\ \hline
0.79 & 0.989 & ~Kids, you are the BEST, the BEST the BEST, the BEST FUTURE! \\ \hline
0.83 & 0.989 & -Kids, this was the BEST DAY OF YOUR LIFE! \\ \hline
0.86 & 0.990 & -Kids, today we’re going to have the BEST DAY OF OUR LIFE. \\ \hline
0.89 & 0.991 & -Kids, today we’re going to have the BEST DAY OF OUR LIFE!! \\ \hline
0.93 & 0.991 & -Kids, we are so happy to have met you. We love you both!! \\ \hline
0.95 & 0.991 & Kids, we are so happy to have met you! We love you both!! \\ \hline
1.00 & 0.992 & Kids we are so excited that you came into our lives today! Thank you for making our day a little brighter. \\ \hline
\end{tabular}
\caption{\label{table:fullpareto}Full pareto front of a representative run of sentiment modification for the Homer Simpson quote.}
\end{table}
\end{center}

For the Gore Vidal quote, ``Whenever a friend succeeds, a little something in me dies.'' a representative Pareto front (from LMX-Near) is shown in Table~\ref{table:fullpareto2}. 

\begin{center}
\begin{table}
\begin{tabular}{|c|c|p{5cm}|}
\hline
\textbf{Distance} & \textbf{Positivity} & \textbf{Sentence} \\ \hline \hline
0.00 & 0.389 & Whenever a friend succeeds, a little something in me dies. \\ \hline
0.26 & 0.662 & Whenever a friend succeeds, the little things in me die. \\ \hline
0.30 & 0.796 & When a friend succeeds, the little things in me die. \\ \hline
0.31 & 0.892 & When a friend succeeds, I die a little. \\ \hline
0.40 & 0.948 & When a friend succeeds, a little thing in me lives. \\ \hline
0.52 & 0.949 & if a friend succeeds, a big thing in me lives. \\ \hline
0.56 & 0.977 & If a friend succeeds, a great thing comes out of me. \\ \hline
0.59 & 0.985 & If a friend succeeds, that’s the most awesome thing that’s happened to me. \\ \hline
0.63 & 0.985 & If a friend succeeds, I get an exciting feeling in my life, because of them. \\ \hline
0.66 & 0.986 & If a friend succeeds, my friends have the most exciting feeling in my life, because of them. \\ \hline
0.69 & 0.988 & If a friend succeeds, I have the most excitement in my life, because of them. \\ \hline
0.82 & 0.990 & And I'm happy for this friend--I'm happy for this friend. \\ \hline
0.88 & 0.991 & and I'm so happy that I found my new best friend, I'm so happy that I found my new best friend, \\ \hline
\end{tabular}
\caption{\label{table:fullpareto2}Full pareto front of a representative run of sentiment modification for the Gore Vidal quote.}
\end{table}
\end{center}

For the Woody Allen quote, ``Life is divided into the horrible and the miserable'', a representative Pareto front (from LMX-Near) is shown in Table~\ref{table:fullpareto3}. 

\begin{center}
\begin{table}
\begin{tabular}{|c|c|p{5cm}|}
\hline
\textbf{Distance} & \textbf{Positivity} & \textbf{Sentence} \\ \hline \hline
0.00 & 0.013 & Life is divided into the horrible and the miserable. \\ \hline
0.20 & 0.347 & Life is, not divided into the horrible and the miserable. \\ \hline
0.24 & 0.460 & Life is, not divided into the horrible or the miserable. \\ \hline
0.30 & 0.651 & For you are the Life, not divided into the horrible, the miserable. \\ \hline
0.34 & 0.704 & You are the Life, not divided into the horrible or the miserable. \\ \hline
0.41 & 0.751 & This is the eternal life, not divided into the horrible or the miserable. \\ \hline
0.46 & 0.789 & You are the eternal life, not divided into the horrible or the miserable. \\ \hline
0.49 & 0.893 & You are the beautiful life, not divided into the horrible or the miserable. \\ \hline
0.51 & 0.902 & You will see the beautiful life, not divided into the horrible or the miserable. \\ \hline
0.53 & 0.910 & You will be the beautiful life, not divided into the horrible or the miserable. \\ \hline
0.73 & 0.970 & Happiness is the way to live. \\ \hline
0.79 & 0.987 & Happiness is the way to live. And I'm very happy with the way that I live. \\ \hline
0.83 & 0.988 & My life is wonderful, I'm very happy with the life. \\ \hline
0.84 & 0.990 & We will live in the glorious happiness. And it is really good, it is really good. And I'm very happy with the life that I have. \\ \hline
0.92 & 0.991 & And I'm very happy with the life that I have. And I can't wait to see the next one. \\ \hline
\end{tabular}
\caption{\label{table:fullpareto3}Full pareto front of a representative run of sentiment modification for the Woody Allen quote.}
\end{table}
\end{center}

\subsection{Evolving towards Negative Sentiment}
\label{appendix:negative}

We also did some initial experiments targeting the negative sentiment class instead of the positive one, i.e.\ taking positive quotes and turning them negative. As in the experiments in the paper, LMX is able to successfully evolve modifications to quotes that achieve high negativity. However, it often does so by evoking vulgar language or dark situations (e.g.\ the death of loved ones, or depressive thoughts about hate). 

It does often make resigned versions of common inspirational quotes; e.g. one negative version of ``Be the change that you wish to see in the world,'' it produces is ``you can't be the change you want to see in the world.'' From the same run, the most negative sentence on the Pareto front is: ``you are the world's worst failure, you have not had good news for the last six months, and you will never find a way to make it up.'' From the inspirational quote ``When the sun is shining I can do anything; no mountain is too high, no trouble too difficult to overcome,'' it creates a dreary version: ``The earth and the mountains beat me hard, the winds blow heavily, the weather is bitter and cold; I cannot do anything.''

While the results are not always pleasant, these preliminary experiments highlight that by using a different classification label (or potentially a different model that recognizes different properties of text altogether), it is possible to use LMX for style-transfer of possibly many other styles.

\section{Image Generation Experimental Details}
\label{appendix:imagegen}

The image generation experiment used Stable Diffusion v1.4 as the text-to-image model for generating images from evolved prompts; specifically, the 16-bit weight variant (fp16), run from the HuggingFace \texttt{diffusers} library.\footnote{\url{https://github.com/huggingface/diffusers}} Images were generated at the default $512 \times 512$ resolution, and generation was run for 10 diffusion steps per image. While the default number of steps is normally 50, performance was valued over image fidelty. We left the default NSFW filter enabled, which produces a black image when triggered.

Image fitness functions were computed using 8-bit integer RGB images; the maximum fitness for the excess-red, excess-blue, and excess-green fitness functions is therefore $512 \cdot 512 \cdot 255 = 66{,}846{,}720$, which would be the fitness of a monochromatic image of the target color. The plot in Figure~\ref{fig:sdx_results} is normalized with 1.0 set to this maximum fitness.

Pythia-deduped 2.8b was used as the LLM. This is from the same Pythia model series discussed in Appendix \ref{appendix:binary}. Up to 75 tokens were sampled from the LLM for each LMX-generated prompt, to stay under Stable Diffusion's limit of 77 tokens in a prompt (Pythia and Stable Diffusion have slightly different tokenizers).

The GA loop used for these experiments was identical to the one from the symbolic regression experiments.
The same tournament selection process was used, as well as the same 0.05 probability of drawing a new human-written prompt from the initial dataset instead of performing LMX (0.95 probability of generating a new prompt through LMX).
Four parents were used as prompts to the LLM to produce each LMX-generated child.
The number of parents was not tuned for this problem, but chosen based on the results in Figure~\ref{fig:binary_1}a.
Each parent was placed in a paragraph by itself prefixed by ``Prompt: ''. The list of parents ended with an open ``Prompt:'' to request that a child be generated.
On an NVIDIA GeForce RTX 3090, with a population size of 50, each generation took about 4 minutes of wall-clock time.

\section{Python Sodaracers Experimental Details} \label{appendix:python}

Experiments for the Sodaracers domain were carried out using Salesforce's CodeGen suite of language models \cite{nijkamp2022codegen}, using the 350M, 2B, and 6B sizes in their `mono' variant. The `mono' models were first pre-trained on natural language, before being fine-tuned on a large dataset of code in many languages, before finally being fine-tuned on a dataset of Python only code. All model sampling was done with top p $= 0.95$, temperature $= 0.85$, and with a maximum generation length (in addition to the prompt) of 512 tokens. Evolutionary runs, as described in Section~\ref{sec:sodaracer}, took up to 30 hours (at 6B scale) to run on a single Nvidia A100 40GB GPU, while smaller models were significantly quicker. Use of Nvidia's Triton Inference Server has the potential to speed up sampling from these language models by up to an order of magnitude.

The seven Sodaracers used as our seed programs are described in the appendix of Lehman et al.~\cite{lehman2023evolution}: the square, radial, wheel, runner, galloper, CPPN-Fixed, and CPPN-Mutable programs (CPPN stands for Compositional Pattern-Producing Network \cite{stanley2007compositional}).

The Sodaracers were evaluated in a Python simulation of the Sodarace domain \cite{szerlip2013indirectly} written in Box2D (from the Open ELM project \cite{bradley2023openelm}). The fitness function was measured as the horizontal distance travelled by an instantiated robot after 1 second of simulation time; this abbreviated evaluation time is enough to show meaningful locomotion and was chosen to demonstrate meaningful evolution within our computational constraints.

\begin{figure} 
    \centering
    \includegraphics[width=0.4\columnwidth]{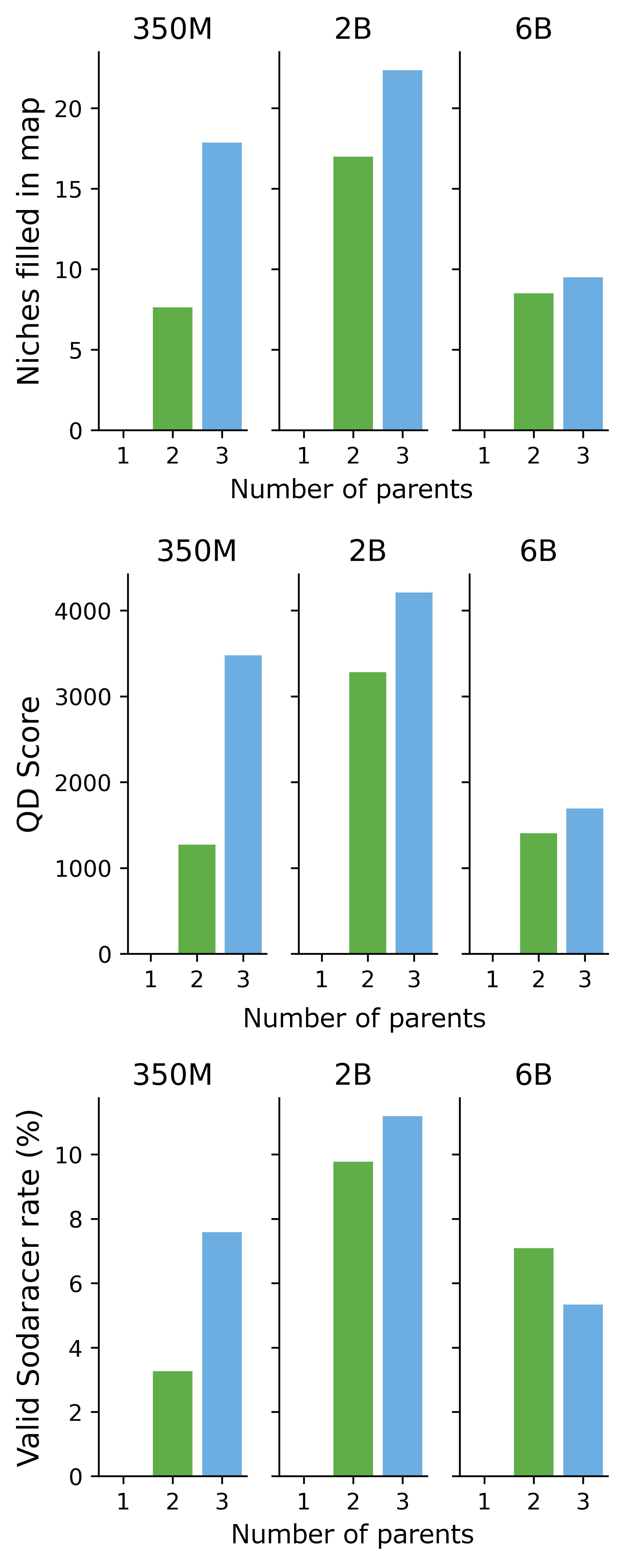}
    \caption{Results from initial experiments in the Sodaracers domain, using a prompt that concatenates parent programs together and forces the correct function signature as described in Appendix~\ref{appendix:python}, for varying numbers of parents in the language model prompt and across language model scale. This experiment consists of 1000 steps of initialization of Sodaracers from the initial seed programs with no evolution. (top) Number of niches filled in MAP-Elites. (center) Quality-Diversity scores (bottom) Validation rate (\%) for the generated Sodaracers. Higher numbers of parents nearly always increases performance in this setting, and the 2B model performs the best (interestingly, the 6B model underperforms; the reason is unclear, but suggests follow-up work to better understand what LLMs best fit a given task).}
    \label{fig:sodaracers_unforced}
\end{figure}

As observed in prior work with few-shot prompting of language models \cite{lu2021fantastically}, we noticed that success rates (the percentage of generations which resulted in valid Sodaracers) varied dramatically with the order of parent functions in the prompt, sometimes by over 50\%. To control for this we either averaged our results over every possible permutation of parents or (in long evolution runs) randomly selected from the set of permutations for each sample.

The main experiments in the paper, described in Section~\ref{sec:sodaracer}, prompt the language model with a concatenation of the seed functions, any necessary Python import statements, and the line \texttt{def make\_walker():} was appended to the end, in order to `force' the language model to complete a function with this signature.

We also experimented with removing this signature from the end, which produces slightly worse results, particularly in terms of the validation rate. For single-seed prompt mutation, all generations failed to validate, while for LMX with two or three parents the validation rate fell by ~15\% compared with the main prompt.

In addition, we cursorly investigated adding an `instruction' to the end of the LMX prompt, such as variants of `Combine the starting programs above to create a new program'. This provides some minimal domain customisation to the language model, and is reminiscent of prior work demonstrating that fine-tuning language models on tasks described as instructions can dramatically improve performance on unseen tasks \cite{sanh2021multitask, wei2021finetuned}. While promising, we did not further pursue this direction, but believe such instruction prompting is an intriguing direction for future work with instruction-finetuned language models, which may offer improved quality and diversity of evolved programs or strings if prompted in a way compatible with their training data (as discussed in Section~\ref{sec:discussion}).

\end{document}